\pdfoutput=1

\documentclass[11pt]{article}

\usepackage[final]{acl}

\usepackage{times}
\usepackage{latexsym}

\usepackage[T1]{fontenc}

\usepackage[utf8]{inputenc}

\usepackage{microtype}

\usepackage{inconsolata}

\usepackage{graphicx}

\title{\dataset: Evaluating Long-Context \& Long-Form Retrieval-Augmented Generation with Key Point Recall}

\author{Zehan Qi$^{1*}$, Rongwu Xu$^{1*}$, \\
\bf Zhijiang Guo$^{2\dag}$, Cunxiang Wang$^{3}$, Hao Zhang$^{4}$, Wei Xu$^{1\dag}$\\
  $^{1}$Tsinghua University $^{2}$University of Cambridge \\
  $^{3}$Westlake University, $^{4}$Nanyang Technological University\\
  \texttt{\{qzh23, xrw22\}@mails.tsinghua.edu.cn} \\
$^{*}$ Equal contribution, $^{\dag}$ Corresponding authors}

\usepackage{hyperref}
\usepackage[all]{hypcap}

\usepackage{tcolorbox}
\usepackage{amsmath}
\usepackage{amssymb}

\usepackage{booktabs}
\usepackage{threeparttable}
\usepackage{tabularx}
\usepackage{xspace}
\usepackage{soul}
\usepackage{xcolor}
\usepackage{subfig}

\usepackage{bm} %

\usepackage{algorithm}
\usepackage{algorithmic}

\usepackage{multirow}

\newcommand{\ie}{{\em i.e.}}

\newcommand{\wrt}{{\em w.r.t.}}

\newenvironment{packeditemize}{
\begin{list}{$\bullet$}{
\setlength{\labelwidth}{8pt}
\setlength{\itemsep}{0pt}
\setlength{\leftmargin}{\labelwidth}
\addtolength{\leftmargin}{\labelsep}
\setlength{\parindent}{0pt}
\setlength{\listparindent}{\parindent}
\setlength{\parsep}{3pt}
\setlength{\topsep}{3pt}}}{\end{list}}

\usepackage{pifont}

\tcbset{
    mybox/.style={
        fontupper=\linespread{.9}\selectfont,
        top=0pt, %
        bottom=0pt %
    }
}
\newcommand{\eli}{\textsc{ELI5}}

\newcommand{\dataset}{\textsc{Long$^2$RAG}\xspace}

\newcommand{\metric}{\textit{KPR}\xspace}

\newcommand{\kpp}{\textit{KPP}\xspace}

\newcommand{\kpf}{\textit{KPF}\xspace}

\newcommand{\llmdataset}{$\text{LLM}_{\texttt{dataset}}$}

\newcommand{\llmevaluator}{$\text{LLM}_{\texttt{evaluate}}$}

\definecolor{mygreen}{RGB}{11,141,10}
\definecolor{myred}{RGB}{222,52,57}
\definecolor{myorange}{RGB}{244,140,60}
\definecolor{myblue}{RGB}{70,130,180}
\definecolor{mydeepblue}{RGB}{65,105,225}
\definecolor{myviolet}{RGB}{97,0,138}
\definecolor{myburgundy}{RGB}{110,10,30}
\definecolor{myblue2}{RGB}{0,105,148}
\definecolor{iceblue}{RGB}{173, 216, 230}
\definecolor{puregreen}{RGB}{0, 218, 0}
\definecolor{graygreen}{RGB}{74,113,106}

\definecolor{wingreen}{rgb}{0,0.45,0.24}
\definecolor{losered}{rgb}{1.0,0.1,0.24}

\definecolor{lightcoral}{rgb}{0.97, 0.36, 0.46}
\definecolor{lightyellow}{rgb}{0.98, 0.7, 0}
\definecolor{harvestgold}{rgb}{0.85, 0.57, 0.0}
\definecolor{brightlavender}{rgb}{0.75, 0.58, 0.89}
\definecolor{capri}{rgb}{0.0, 0.75, 1.0}
\definecolor{carminepink}{rgb}{0.92, 0.3, 0.26}
\definecolor{celadon}{rgb}{0.67, 0.88, 0.69}
\definecolor{darkpastelgreen}{rgb}{0.01, 0.75, 0.24}

\definecolor{grayhighlight}{RGB}{250,250,227}

\begin{document}
\maketitle

\begin{abstract}
Retrieval-augmented generation (RAG) is a promising approach to address the limitations of fixed knowledge in large language models (LLMs). However, current benchmarks for evaluating RAG systems suffer from two key deficiencies: (1) they fail to adequately measure LLMs' capability in handling \emph{long-context retrieval} due to a lack of datasets that reflect the characteristics of retrieved documents, and (2) they lack a comprehensive evaluation method for assessing LLMs' ability to generate \emph{long-form responses} that effectively exploits retrieved information. To address these shortcomings, we introduce the \dataset benchmark and the Key Point Recall (\metric) metric. \dataset comprises 280 questions spanning 10 domains and across 8 question categories, each associated with 5 retrieved documents with an average length of 2,444 words. \metric evaluates the extent to which LLMs incorporate key points extracted from the retrieved documents into their generated responses, providing a more nuanced assessment of their ability to exploit retrieved information. Our dataset and scripts are available at \href{https://github.com/QZH-777/longrag}{this url}.
\end{abstract}

\section{Introduction}
\label{sec:intro}

\begin{figure}[ht]
    \centering
    \includegraphics[width=\linewidth]{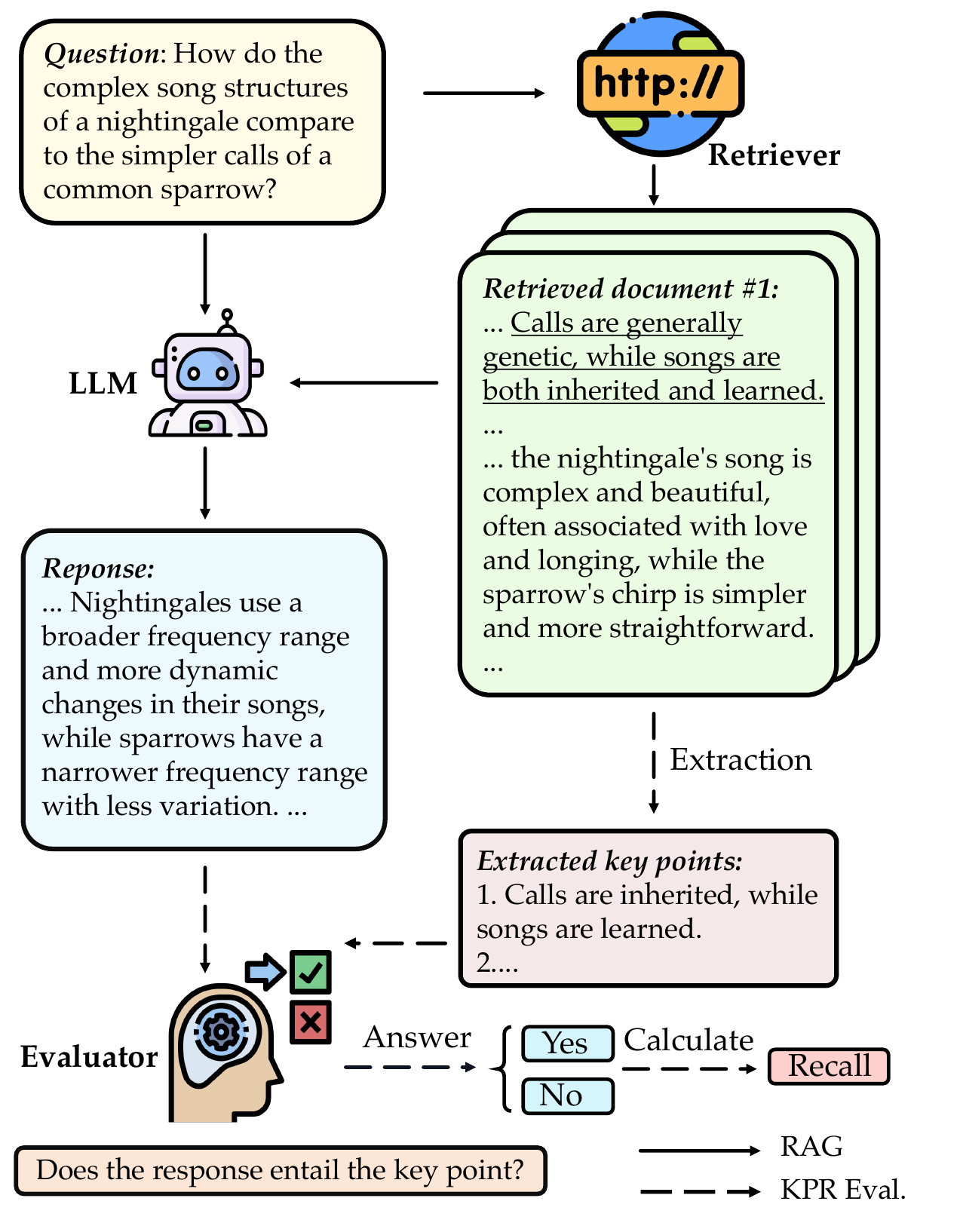}
    \caption{Illustration of the RAG and evaluation pipelines using \metric{}. We first extract the key points from the retrieved documents and compute the recall of these points in the response of the LLM with the help of an \emph{Evaluator} (possibly another LLM), thereby enabling the evaluation of the response quality.}
    \vspace{-1em}
    \label{fig:intro}
\end{figure}

Large language models (LLMs;~\citealt{touvron2023llama, openai2023gpt4, Mixtral2024}) have demonstrated remarkable capabilities across a wide range of tasks.
However, the fixed and finite nature of the knowledge embedded in LLMs presents limitations~\cite{he2022rethinking, xu2024knowledge}.
Retrieval-augmented generation (RAG), which incorporates external knowledge through search engines, represents a promising avenue for addressing this constraint~\cite{borgeaud2022improving, gao2023retrieval}. 

Recent efforts in \emph{long-context LLMs}~\cite{xiong2023effective, liu2024lost} leads to the packing of complete document content into LLMs to prevent information loss~\cite{xu2023retrieval, gao2023retrieval}. This also challenges LLM's capability to handle long contexts. Several benchmarks are designed for evaluating long-context understanding~\cite{xu2023retrieval, shaham2023zeroscrolls, liu2024lost}. Nevertheless, none of them consider the characteristics of low signal-to-noise ratio and dispersed information distribution in retrieved documents of RAG, leading to \emph{input-side deficiency}. Existing benchmarks fail to adequately measure LLMs' capability in handling \emph{\textbf{long}-context RAG}.

Meanwhile, current assessment of LLMs within the realm of RAG mainly focuses on short answers~\cite{chen2024benchmarking, zhang2024retrievalqa}, leaving a significant gap in evaluating LLMs' proficiency in generating \emph{\textbf{long}-form responses with RAG}. This gap stems from the lack of a comprehensive evaluation method.
For text generation, early automated metrics such as surface form matching~\citep{lin2004, BanerjeeL05} and semantic representation comparison~\citep{zhang2019bertscore, YuanNL21}, face challenges with long-form content due to their inability to handle the diversity of potential outputs~\citep{celikyilmaz2020evaluation, krishna2021hurdles}. 
Recent studies have explored the utility of LLMs for evaluation~\cite{chiang-lee-2023-large, liu2303g}. However, these methods don't consider the retrieved documents within RAG. 
While some studies propose automated metrics for evaluating long-form generation in RAG~\cite{es2023ragas, saad2023ares}, these metrics primarily focus on faithfulness, \ie{}, whether the generated text is grounded in the retrieved documents.
Therefore, no automated method exists for evaluating LLMs' exploitation of retrieved documents, representing a \emph{output-side deficiency} of current RAG benchmarks.

To bridge this gap, we introduce \dataset, comprising 280 questions spanning 10 distinct domains and encompassing 8 question categories. Each question is associated with 5 retrieved documents, with an average length of 2,444 words per document. \dataset is collected with extensive care and offers several benefits. The questions posed in \dataset are both \emph{intricate} and \emph{practical}, requiring a comprehensive response. 
Furthermore, \dataset is carefully designed to \emph{minimize the risk of data contamination}. To mirror the low signal-to-noise ratio and other characteristics prevalent in real-world scenarios, the associated documents originate from \emph{authentic retrieval procedures}. 
To tackle the output-side deficiency, we propose \metric. 
As depicted in~\autoref{fig:intro}, for each question, we automatically extract key points from the associated retrieved documents that directly contribute to answering the question. Subsequently, we evaluate the extent to which these key points are incorporated into the model's generated response, thereby assessing the effectiveness of LLMs in leveraging retrieved documents.

With \dataset and \metric, we extensively evaluate 9 state-of-the-art LLMs. 
We summarize our findings as follows:
\begin{packeditemize}
    \item Closed-source LLMs represented by GPT-4o are more capable than open-source models, with the smaller open-source model (Phi-3-mini) being able to outperform the larger one of 72B (Qwen2).
    \item The model's capabilities show an overall decreasing trend as the input documents grow.
    \item The standard RAG procedure, \ie, the truncation on retrieved documents, leads to a loss of information, resulting in weaker performance than RAG under long context.
\end{packeditemize}
We hope \dataset can facilitate the understanding of long-context RAG systems
from multiple dimensions and facilitate the development of LLMs in exploiting retrieved information.

\begin{figure*}[ht]
    \centering
    \includegraphics[width=\linewidth]{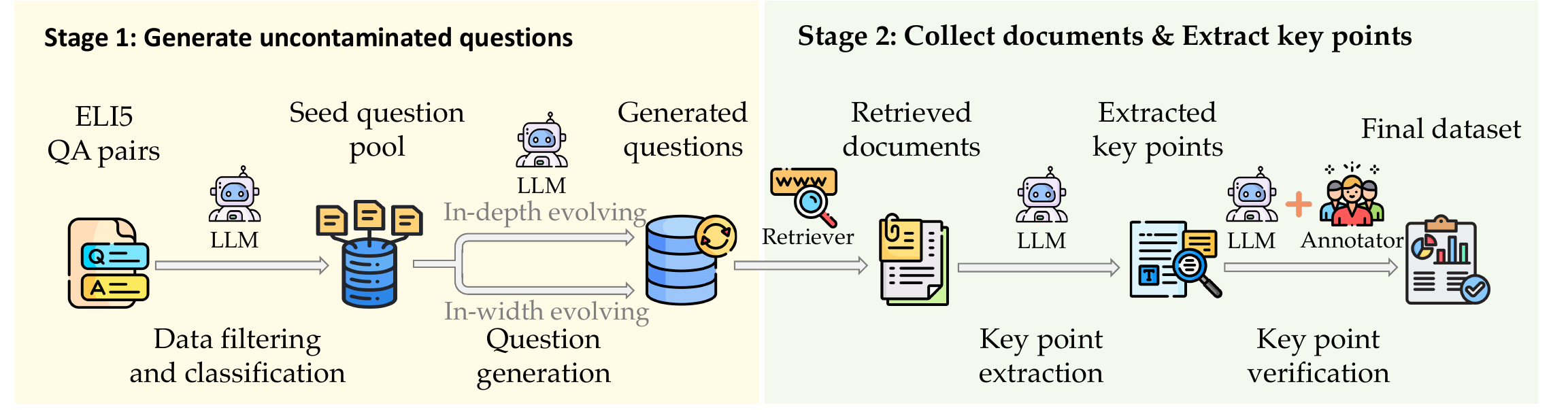}
    \caption{Overview of our dataset construction pipeline. The process comprises \emph{two} main stages. In the first stage, we aim to generate uncontaminated questions by employing an LLM to filter questions from ELI5 and construct a seed question pool. By using two evolving techniques, new questions are generated. 
    In the second stage, a search engine is utilized to procure documents for the RAG pipeline, where the key points are extracted automatically afterward. We finally employ a human-LLM collaborated verification task that result in our final dataset.}
    \label{fig:pipeline}
    \vspace{-1em}
\end{figure*}

\section{Related Works}
\label{sec:related}

\subsection{RAG Benchmarks}

Recent efforts in benchmarking RAG have primarily focused on two distinct evaluation objectives: retrieval and generation.
Research on the \emph{\textbf{retrieval}} aspect aims to assess the quality of the retrieved documents, considering factors such as retrieval relevance and timing~\cite{lyu2024crud, es2023ragas, saad2023ares}. Another line of work, where our study falls into, is concerned with the \emph{\textbf{generation}} process~\cite{chen2024benchmarking, zhang2024retrievalqa, stolfo2024groundedness}, of which can also be categorized into short-form and long-form evaluation. In the former scenario, a succinct reference answer exists, and assessments predominantly rely on conventional metrics such as exact match (EM) and F1 score~\cite{chen2024benchmarking, zhang2024retrievalqa}. 
However, this line of evaluation neglects the fact that people regularly use RAG for generation in real-world applications. For long-form evaluations,
\citet{gao2023enabling} evaluate the citation relevance during generation, while we evaluate the model's ability to identify key points, regardless of citations.
Several recent studies evaluate the \emph{precision}\footnote{Whether the claims within the generated text are grounded by retrieved documents.} of model-generated texts~\cite{stolfo2024groundedness, es2023ragas, saad2023ares}. Our research, on the contrary, adopts a \emph{\textbf{recall}} perspective, assessing how well the generation captures key information present in retrieved documents.
While a related work, namely CRUD~\cite{lyu2024crud}, also employs recall-like measurements, it focuses on utilizing one gold reference, which is not obtained by retrieval, to evaluate the generation. In contrast, our research is primarily oriented toward evaluating a model's ability to utilize useful information from \emph{multiple retrieved documents}.

\subsection{Text Generation Evaluation}
We classify long-form text generation evaluation into reference-based evaluation where gold answers are required and reference-free evaluation~\cite{wei2024long, lyu2024crud}. 
In the former case, methods employed to evaluate the similarity between the generation and the gold answer~\cite{fan2019eli5, chiang-lee-2023-large} face challenges in acknowledging the legitimate range of potential appropriate answers~\cite{krishna2021hurdles, celikyilmaz2020evaluation}. 
In addition, the evaluation outcomes may not align well with human judgments~\cite{xu2023critical}. 
Motivated by these drawbacks, reference-free evaluation has attracted considerable interest, with some endeavors concentrating on assessing the coherence and relevance of the generation to specified \emph{questions}~\cite{ fabbri2021qafacteval, krishna-etal-2022-rankgen, xu2023critical, xu2024debateqa}. 
Another line of literature explores the factuality of model generation by leveraging external \emph{knowledge bases}~\cite{stelmakh-etal-2022-asqa, min-etal-2023-factscore, wei2024long}. 
The work most closely related to ours is ProxyQA~\cite{tan2024proxyqa}, which evaluates long-form generation through expert-designed \emph{proxy questions}. We distinguish our approach by (1) focusing on the utilization of external knowledge in the RAG setup, (2) proposing an alternative \emph{key point recall} measurement, and (3) largely reducing the need for human expert involvement in constructing key points.

\section{Dataset Construction}
\label{sec:dataset}
In this section, we introduce the process of constructing \dataset. 
Overall, We leverage an automated pipeline\footnote{We use GPT-4-Turbo as the \llmdataset{}.} as illustrated in~\autoref{fig:pipeline}. The generated dataset then went through human-LLM collaborated verification to ensure the quality.

To start with, we aim to create \emph{questions} with the following properties:
\begin{packeditemize}
    \item The questions are complex and \emph{cannot} be easily answered by LLMs utilizing their parametric knowledge.
    \item The questions are practical and require a long-form answer.
    \item The questions are uncontaminated and thus less likely to be memorized by the LLMs.
\end{packeditemize}
Having prepared the questions, we utilize an automated approach to extract the \emph{key points} from \emph{retrieved documents} (obtained by leveraging a search engine), which will serve as the basis of our later evaluation.
Finally, \dataset{} includes domain- and characteristic-diverse questions, paired with real-world retrieved documents and automatically extracted and human-verified key points \wrt{} to each question.

\subsection{Question Generation}
To ensure the practicality of the question, our starting point is \eli{}~\cite{fan2019eli5}, a dataset collecting questions asked by users and corresponding answers from Reddit. However, questions in \eli{} face a potential risk of data contamination~\cite{li2023open, golchin2023time}, \ie{}, being utilized as training material for LLMs. 
To address this drawback, we apply the Evol-Instruct~\cite{xu2023wizardlm} method to further evolve existing questions in \eli{} to generate \emph{fresh} ones. 

\begin{figure*}[ht]
    \centering
    \includegraphics[width=\linewidth]{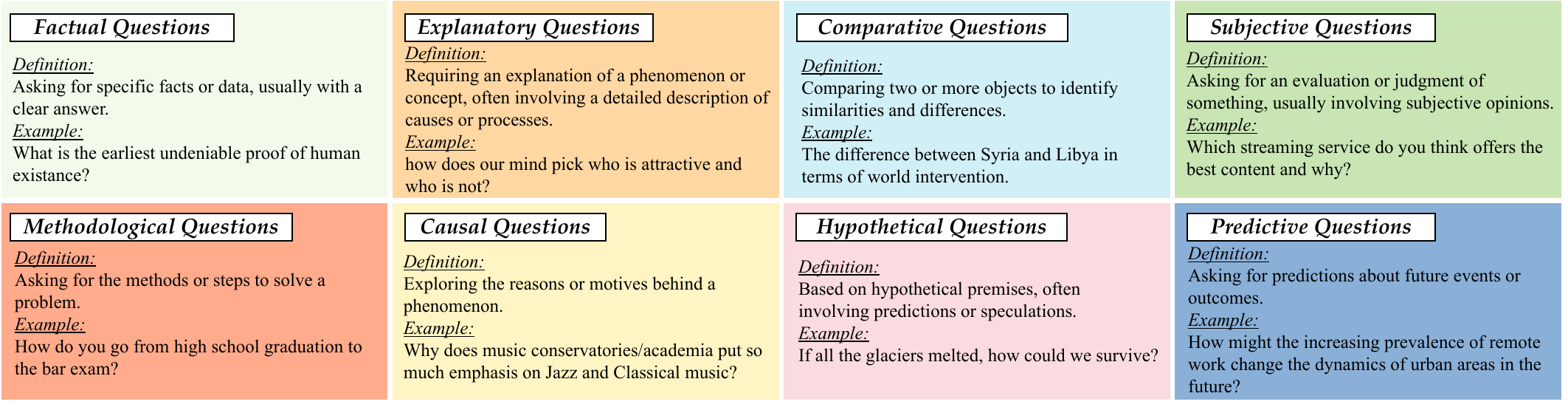}
    \caption{Detailed information about our defined question categories, including definitions and examples.}
    \label{fig:question_category}
\end{figure*}

To ensure consistency between questions acquired by Evolve-instruct and those sourced from \eli{}, we filter questions in \eli{} and create a seed question pool, which can be used to control the newly generated questions. Initially, questions within \eli{} are ranked based on the length of their corresponding answers, with the top 3,000 selected for further scrutiny. This subset then undergoes a filtering process utilizing an \llmdataset, guided by specific criteria: (1) exclusion of common sense questions, (2) insurance of clarity and no ambiguity, (3) fulfillment of complexity requirements, and (4) solicitation for subjective opinions.
Each question is assessed against the aforementioned criteria, with one point assigned for each criterion met. 
Questions accruing four or more points are retained, resulting in an \emph{initial pool} comprising 1,445 questions. Subsequently, we categorize questions in the initial pool by \llmdataset{} into 8 categories as shown in~\autoref{fig:question_category}. For each category, a manual filtration process is employed to identify the top 12\footnote{For subjective questions, it is insufficient to select 12 questions from the initial pool that meet the standards, we select 7 questions for this category instead.} questions align most closely with the defining attributes of this category, serving as the \emph{seed question}.
Finally, these 91 seed questions serve as our \emph{seed question pool}.

Employing both the in-depth evolving method and the in-width evolving method in Evolve-Instruct, new questions are generated for each category. 
In addition to regulating the category of the generated questions, we also determine the domain in which these questions should be focused. A selection of 10 domains is designated to guide the question-generation process. Following the generation phase, a manual screening process is employed. 
Given the scale of the dataset under consideration, we decide to save 7 questions for each category-domain pair. 
This generation-screening process is repeated until a sufficient number of questions are obtained. So far, a dataset comprising 560\footnote{560 = 8 categories $\times$ 10 domains $\times$ 7 questions.} questions is created.

\subsection{Document Collection}
We adopt a question decomposition strategy leveraging the capabilities of \llmdataset{} to break down each question into several sub-questions to improve the quality of the retrieval process. 
We utilize two APIs, including the \href{https://developers.google.com/custom-search/v1/introduction}{Google search engine} and \href{https://www.langchain.com.cn/modules/agents/tools/examples/google_serper}{Serper search engine} to retrieve pertinent documents for each subquestion. The top 5 documents obtained from each API are saved. Instead of using snippets, the complete content of each document is saved. For all retrieved documents from the subquery, a model\footnote{Bge-Reranker-v2-Gemma is the model we use to rerank all retrieved documents based on their score with the query.} is used to rerank them, with only the top-5 most relevant to the query being retained. 
After this step, each question is accompanied by 5 most relevant real-world retrieved documents.

\subsection{Key Point Extraction}
\label{sec: extraction}
We use an automated pipeline\footnote{Given the extensive length of retrieval documents with an average of 2,453 (as shown in~\autoref{fig:length-distribution}) and demand for expert knowledge, exclusive reliance on human effort for extraction is impractical.} to extract key points from retrieved documents using \llmdataset{}. 
In \dataset{}, we define ``key point'' to be a concise, self-contained piece of information from the source documents that is both necessary and sufficient to formulate a complete and accurate answer to the given question, akin to the \emph{score point in the grading process of an examinee's problem solution}.
The detailed instructions for extracting the key points are listed in~\autoref{sec:prompts-dataset}.
To ensure comprehensive coverage, we conduct multiple rounds of extraction, each time prompting GPT-4 to consider previously overlooked information.
All the extracted points for each question are then de-duplicated and aggregated by using the same LLM, leading to a more general expression of these key points. To ensure the aggregated points are complete and disjoint, the LLM also outputs the original points corresponding to each aggregated point. We check the aggregation results by seeing if an original point is in more than one aggregated point or not in any of them.
Ultimately, this process yields a dataset comprising 28,611 key points. 

We manually verify the effectiveness of this process on a subset of 20 randomly-sampled questions. For the documents associated with 10 randomly selected questions, we compare the model-extracted key points against human-identified key points.
We calculate the recall of the model-extracted key points against the human-identified ones, achieving a 98.5\% recall.

\noindent \textbf{Human-LLM collaborated verification of the key points.}
We conduct a human-LLM collaborated verification to ensure the quality of the extracted key points.
After sampling and observation, we find that most of the extracted points are not key points but only relevant to the question, with about 30\% of the points meeting our definition. 
To reduce the cost of manual annotation, we apply the scoring function in AutoDS~\cite{zhang2024autonomous} to coarsely filter these points (see Appendix~\autoref{sec: score_func} for details). Using this score function, we calculate a score for each point to indicate its significance. 
We set 0.7 as the threshold, which demonstrates an 8.7\% false positive rate by manually annotating a set of 20 questions with 1,051 key points. Details are deferred to Appendix~\autoref{sec: score_func}. 
This results in a final 8,457 points remaining. Subsequently, we recruited 4 annotators to annotate the remaining points. They are encouraged to use the search engine to resolve any concepts that remain unclear during the annotation process. Their work is then subject to a spot-check to ensure that they adhere to the established guidelines. 

Constrained by the annotation quality, we decided to use \textbf{2 out of the 4} annotators' annotation results (280 out of the 560 questions, the distribution of domains and question categories of the 280 results are shown in Appendix~\autoref{sec: distribution}).
The 2 annotators achieve an inter-annotator agreement (IAA) of 0.56, indicating moderate agreement. The IAA is assessed by the free-marginal multi-rater kappa~\cite{randolph2005free}.

\noindent \textbf{Final dataset.} 
Initially, our automatic pipeline identified 3,651 key points linked to 280 questions. Following the human annotation process, we excluded points if they were not grounded by the document or did not meet the criteria for being considered a key point. Consequently, we retained 2,055 key points, which represents a 56.3\% retention rate of the original points extracted by our automatic pipeline. While the dataset size of 280 questions might seem limited for an eight-category benchmark, it is comparable in scale to similar manually verified datasets like ProxyQA\citep{tan2024proxyqa}, which has 100 questions across 9 categories. We emphasize quality over quantity, selecting 280 high-quality questions from an initial 560 to ensure accuracy and reliability. Unlike other datasets that rely on scraped or extracted data, our manual annotation process minimizes contamination and inconsistencies, making it a robust resource for evaluating long-form text generation models.

\section{\metric{}: a Newly Introduced Evaluation Metric}
\label{sec:dataset}

We propose the evaluation metric termed Key Point Recall (\metric) to evaluate to which extent the model exploits the retrieved documents. Consider a given question, denoted as \( q \), for which \(d^q\) indicates the concatenation of retrieved documents and \( \mathbf{x^q} = [x_1^q, x_2^q, \cdots, x_n^q] \) represents the set of all key points of the question. Let \( y \) denote the response generated by a model $\mathcal{M}$, where \(y = \mathcal{M}(q \Vert d^q)\). We define an evaluation function \( I(x_i^q, y) \) to assess the entailment relationship of key point \( x_i^q \) within the model's generation \( y \):
\begin{equation}
  I(x_i^q, y) = 
  \begin{cases} 
   1 & \text{if } y \text{ entails } x_i^q, \\
   0 & \text{otherwise}.
  \end{cases}
\end{equation}
The evaluation function is implemented with an evaluator \llmevaluator{}, using the prompt detailed in~\autoref{sec: score_func}.
Therefore, given a question dataset \( \mathcal{Q} \), the \metric can be calculated as:
\begin{equation}
\fontsize{10.4}{10.4}\selectfont
    \text{\metric}(\cdot)\!=\!\frac{1}{|Q|}\!\sum_{q\in Q}\!{\frac{\sum_{x \in \mathbf{x^q}}\!{I(x,\!\mathcal{M}(q \Vert d^q))}}{|\mathbf{x^q}|}},
\end{equation}
where \metric calculates the average key point coverage in the question dataset.

\section{Experiment}
\label{sec:experiment}

In this section, we apply \dataset{} and \metric{} to evaluate the performance of state-of-the-art LLMs.

\subsection{Experimental Setup}

\noindent \textbf{Evaluated Models.}
We evaluate a vast array of both commercial API LLMs and open-source LLMs. 
For API LLMs, we select 3 models, including \href{https://openai.com/index/hello-gpt-4o/}{GPT-4o}, GPT-4-Turbo~\cite{openai2023gpt4}, and \href{https://www.anthropic.com/news/claude-3-family}{Claude-3-Sonnet}.
For open-source LLMs, we select 6 models of different sizes considering their parameters. 
We select \href{https://qwenlm.github.io/blog/qwen2/}{Qwen2} models in small ($< 7$B), medium ($\sim 7$B), and large size ($\sim 70$B). We also incorporate popular LLMs including Phi-3-mini-128K~\cite{abdin2024phi} and Mis(x)tral-Instruct (both 7B~\cite{Mistral2023} and \href{https://mistral.ai/news/mixtral-8x22b/}{8*22B} Sparse Mixture of Experts (SMoE)~\cite{Mixtral2024} variants) in the evaluation.

For all models, we employ a straightforward approach to integrate the retrieved documents. We concatenated multiple retrieved documents directly, ensuring that clear separators were placed between each to maintain coherence. Subsequently, we appended a tailored prompt: \emph{``Your answer should incorporate as many important points from the documents as possible that help in answering the question.''}. This prompt was designed to encourage the generation of comprehensive, long-form responses.
We configure all models using greedy decoding.

\noindent \textbf{Selection of the Evaluator.}
In theory, any performance language model could serve as an evaluator \llmevaluator{} for its capability to assess textual entailment. However, when taking into account model performance, inference efficiency, and cost, we chose GPT-4o as our evaluator.

\noindent \textbf{Research Questions.}
In order to conduct an extensive examination of the capabilities of LLMs and to evaluate the efficacy of our proposed \dataset{} benchmark, the following research questions have been formulated to guide our inquiry:

\begin{packeditemize}
    \item \textbf{RQ1:} How well do prevalent LLMs exploit key information in long-context \& long-form RAG? 
    \item \textbf{RQ2:} What is the difference in capabilities for different domains?
    \item \textbf{RQ3:} What is the difference in capabilities for different question categories?
    \item \textbf{RQ4:} How does the length of documents in the context of RAG affect model performance?
    \item \textbf{RQ5:} If a document is truncated on the input side, what is the impact on the performance?
    \item \textbf{RQ6:} Does \metric{} favor longer generation?
    \item \textbf{RQ7:} Does the use of different models as evaluators maintain consistency in \metric{}?
\end{packeditemize}

Next, we will answer these questions one by one.

\subsection{Main Results}

\begin{table*}[ht]
\setlength{\tabcolsep}{1pt} 
\fontsize{8.5}{8.5}\selectfont
\centering
\begin{threeparttable}
\begin{tabular}{lcccccccccc}
\toprule
\textbf{Model} & \textbf{Size} &\textbf{Factual} & \textbf{Explanatory} & \textbf{Comparative} & \textbf{Subjective} & \textbf{Methodological} & \textbf{Causal} & \textbf{Hypothetical} & \textbf{Predictive} & \textbf{\emph{Average}}\\
\midrule
\multicolumn{11}{c}{\textbf{API LLMs}} \\
\midrule
\textbf{GPT-4o} & N/A & \textbf{0.621} & \textbf{0.645} & \textbf{0.658} & \textbf{0.658} & \textbf{0.487} &\textbf{0.559} & \textbf{0.515} & \textbf{0.580} & \textbf{0.579} \\
\textbf{GPT-4-Turbo} & N/A & 0.542 & 0.492 & 0.540 & \underline{0.560} & 0.403 &0.446 & \underline{0.436} & 0.417 & 0.469 \\
\textbf{Claude-3-Sonnet} & N/A & 0.483 & 0.484 & 0.561 & 0.513 & \underline{0.477} &0.437 & 0.394 & \underline{0.537} & \underline{0.477} \\
\midrule
\multicolumn{11}{c}{\textbf{Open-source LLMs}} \\
\midrule
\textbf{Qwen2-Instruct} & 72B & \underline{0.548} & 0.452 & \underline{0.586} & 0.491 & 0.394 &0.417 & 0.414 & 0.392 & 0.449 \\
\textbf{Mixtral-Instruct} & 8*22B\tnote{1} & 0.482 & \underline{0.509} & 0.425 & 0.425 & 0.303 &0.336 & 0.315 & 0.385 & 0.383 \\
\midrule
\textbf{Qwen2-Instruct} & 7B & 0.379 & 0.462 & 0.470 & 0.464 & 0.422 &\underline{0.478} & 0.361 & 0.360 & 0.416 \\
\textbf{Mistral-Instruct} & 7B & 0.509 & 0.426 & 0.456 & 0.474 & 0.290 &0.308 & 0.315 & 0.329 & 0.373\\
\midrule
\textbf{Qwen2-Instruct} & 1.5B & 0.305 & 0.303 & 0.276 & 0.323 & 0.290 &0.225 & 0.243 & 0.186 & 0.262 \\
\textbf{Phi-3-mini-128K} & 3.8B & 0.488 & 0.465 & 0.483 & 0.514 & 0.409 &0.393 & 0.395 & 0.397 & 0.434 \\
\bottomrule
\end{tabular}
\begin{tablenotes}
\fontsize{8}{8}\selectfont
   \item[1.] The activate number of parameters at inference time is \href{https://mistral.ai/news/mixtral-8x22b/}{39B}.
\end{tablenotes}
\end{threeparttable}
\caption{\label{tab:main-result}The \metric{} of various LLMs across different question categories. Overall performance on \dataset{} is reflected by \textbf{\emph{Average}}. \textbf{Highest} and \underline{second-highest} \metric{} are highlighted for each question category.}
\end{table*}

The overall evaluation results of 9 LLMs are shown in~\autoref{tab:main-result}. 
For \textbf{RQ1}, we conclude that: 
\begin{packeditemize}
    \item closed-source API LLMs generally outperform open-source ones, with GPT-4o being the most competent LLM;
    \item overall, in the case of QWen2, an increase in the size of the model results in a corresponding increase in performance;
    \item larger models do not always beat smaller ones, the smaller Phi-3-mini outperforms the 8*22B Mixtral and nearly approaches Qwen2 72B.
\end{packeditemize}

\subsection{Results on Different Domains}

\begin{figure}[ht]
    \centering
    \includegraphics[width=\linewidth]{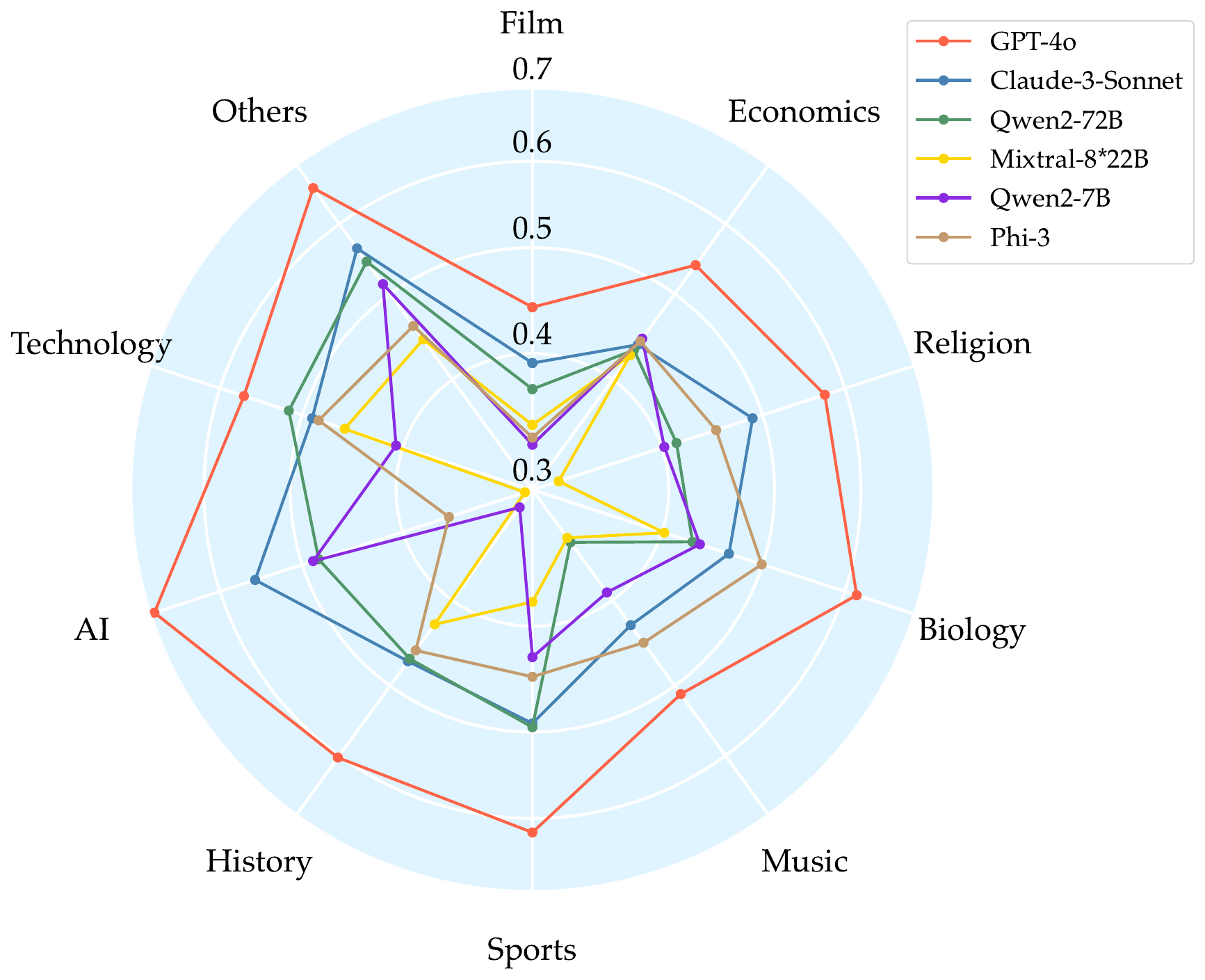}
    \caption{\metric{} of LLMs on different domains.}
    \label{fig:domain-kpr}
\end{figure}

\noindent To answer \textbf{RQ2}, we present a radar plot that compares the model performance across domains in~\autoref{fig:domain-kpr}. 
We discover that GPT-4o outperforms other LLMs (including closed-source ones) across all domains, by a large margin.
Meanwhile, an interesting phenomenon is that all the models we evaluate do not cover the film domain very well. 
In addition, we observe that each model exhibits a degree of specialization in distinct domains. For instance, GPT-4o and Claude-3-Sonnet demonstrate superior performance on problems within the AI domain. In contrast, AI-related problems pose a challenge for Phi-3 and Mixtral, indicating a relative weakness in their domain-specific capabilities.

\subsection{Results on Different Question Categories}

\begin{figure}[ht]
    \centering
    \includegraphics[width=\linewidth]{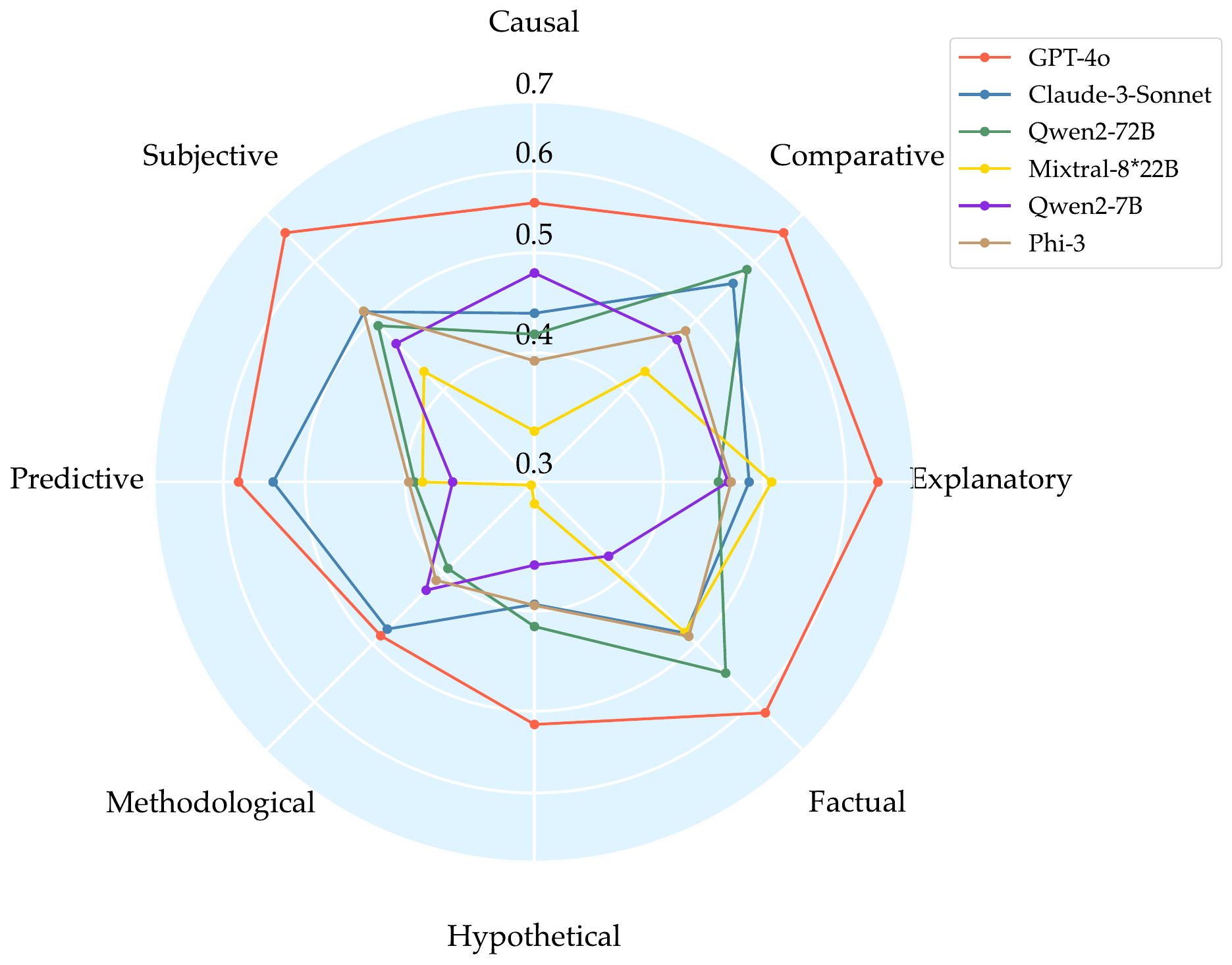}
    \caption{\metric{} of LLMs on different question categories.}
    \label{fig:question-kpr}
\end{figure}

\autoref{fig:question-kpr} depicts the detailed evaluation results within 8 question categories. 
In addressing \textbf{RQ3}, we have identified a pattern analogous to that observed in RQ2. However, the distinction here lies in the types of questions being considered, rather than the domains of expertise.
It is noteworthy that we were pleasantly surprised to discover that nearly every model we evaluated performs exceptionally well on comparative questions, where the models adeptly incorporate important information from both sides of the comparison.

\subsection{Results on Different Documents Length}

\begin{figure}[ht]
    \centering
    \includegraphics[width=\linewidth]{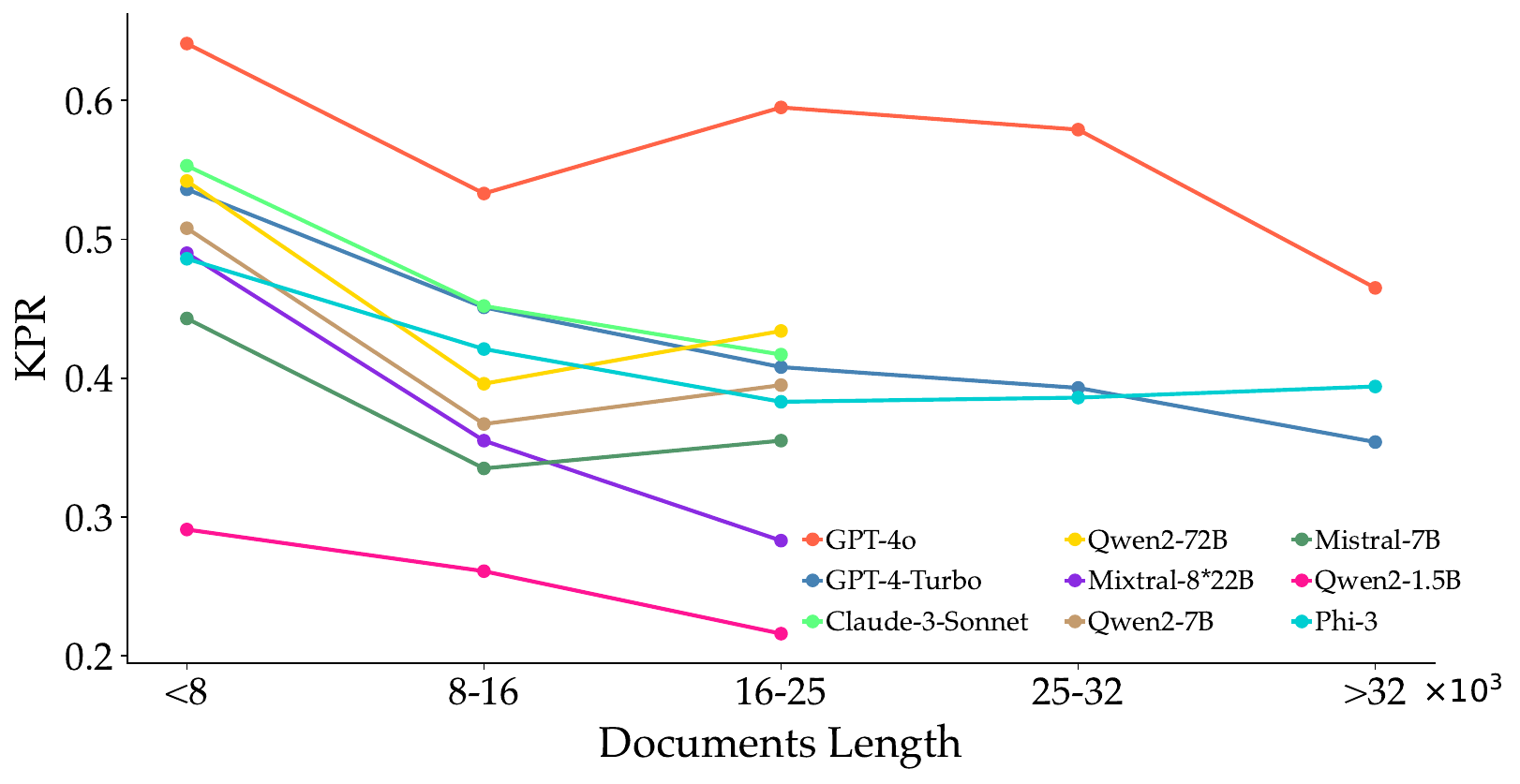}
    \caption{\metric{} of LLMs on different document lengths, where the horizontal coordinate is the length of the packed retrieved documents (in terms of tokens).}
    \label{fig:length-kpr}
\end{figure}

To answer \textbf{RQ4}, we plot the correlation between input documents length after concatenation and \metric{} in~\autoref{fig:length-kpr}.
Upon examining the figure, it is evident that there is a discernible trend: as the length of the input documents increases, the performance tends to deteriorate. This observation is consistent with recent research on long-context comprehension~\cite{liu2024lost, pal2023giraffe}.
Beyond the general trend, we have identified an intriguing pattern: a minor yet noticeable uptick in performance for several models when the generation length increases from 8-16K to 16-25K tokens. This performance rebound could potentially be attributed to the models' increased exposure to and familiarity with this range of data lengths during their training.

\subsection{Results on Other Retrieval Strategy}

\begin{table}[ht]
\setlength{\tabcolsep}{2pt} 
\fontsize{9}{9}\selectfont
\centering
\begin{threeparttable}
\begin{tabular}{lcccccc}
\toprule
\multirow{2}{*}{\textbf{Model}} & \multicolumn{3}{c}{\textbf{Trunk Size}} & \multirow{2}{*}{\textbf{Snippet}} & \multirow{2}{*}{\textbf{Summary}} & \multirow{2}{*}{\textbf{N/A}} \\
\cline{2-4}
\rule{0pt}{2ex} & 512 & 1024 & 2048 \\ 
\midrule
\textbf{GPT-4o} & 0.549 & 0.568 & 0.557 & 0.403 & 0.342 & 0.579 \\
\textbf{Qwen2-72B} & 0.353 & 0.375 & 0.408 & 0.379 & 0.307 & 0.468 \\
\bottomrule
\end{tabular}
\end{threeparttable}
\caption{\label{tab:trunk}The effect of different trunk sizes on LLMs' \metric{}. N/A: No truncation.}
\end{table}

\noindent Remember that for a normal RAG pipeline, documents that are too long need to be truncated~~\cite{luo2024bge, lyu2024crud}. To answer \textbf{RQ5}, we first simulate the truncation process by cutting every document with a granularity of characters~\cite{finardi2024chronicles}, with the result as shown in~\autoref{tab:trunk}. We also apply other processing methods for retrieved documents~\citep{gao2023retrieval} (\ie, using GPT-4o to obtain the snippet and summary of the retrieved documents). Note that we only use other approaches to process the document, while all other operations remain consistent.
We observe that models that accept truncated inputs experience a decline in performance. Moreover, using the snippet of the document will result in a more pronounced performance degradation. The reason for this decline is quite straightforward: these strategies remove essential information from the source documents. Conversely, this observation corroborates the thought that \emph{leveraging long-context to RAG pipelines, contributes to superior generation outcomes}. This is particularly evident when considering the perspective of information exploitation.

\subsection{Results on Different Generation Length}

\begin{figure}[ht]
    \centering
    \includegraphics[width=\linewidth]{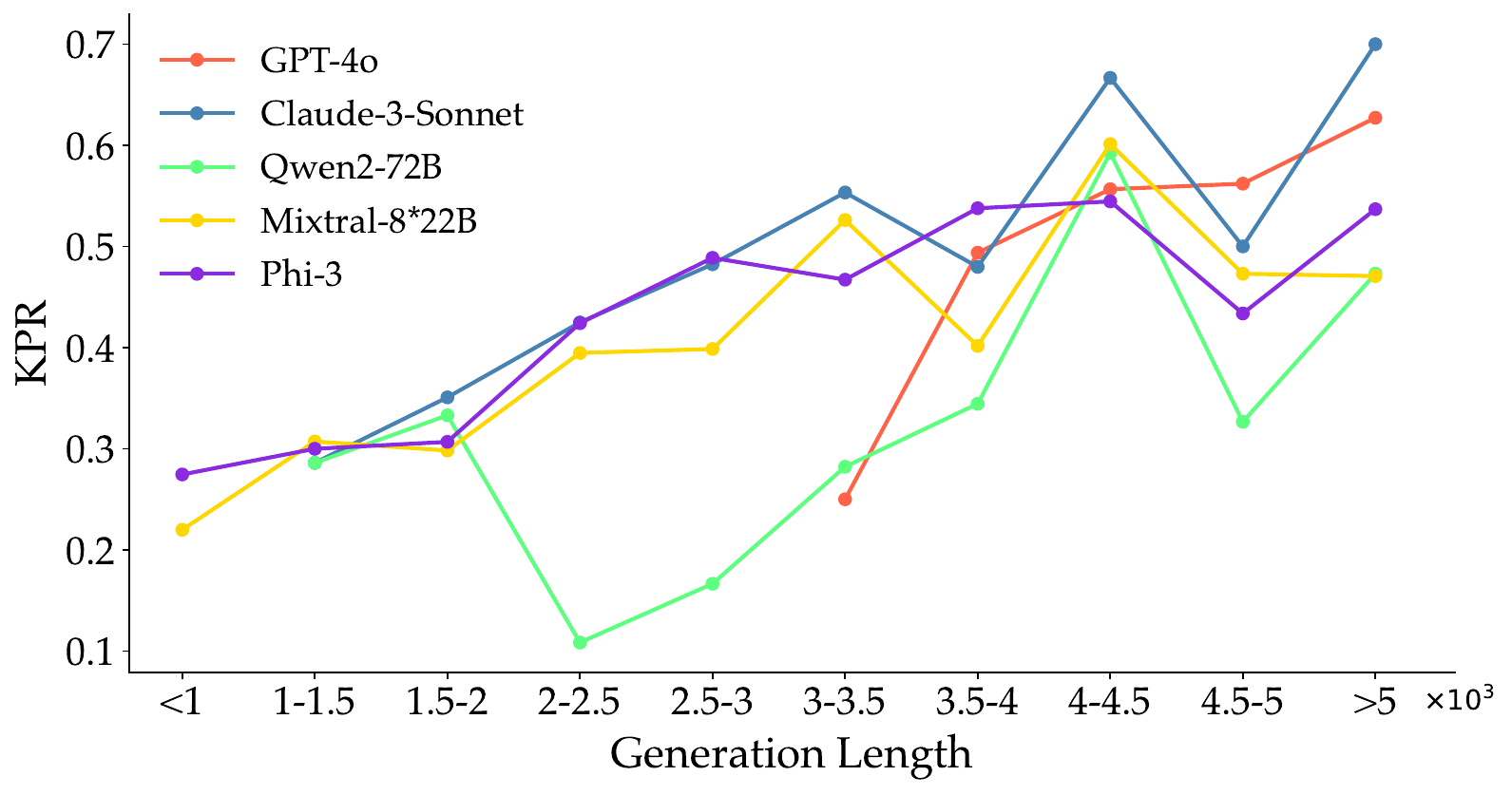}
    \caption{\metric{} of LLMs on different generation lengths (in terms of tokens).}
    \label{fig:gen-length-kpr}
\end{figure}

\begin{figure*}[t]
    \centering
    \includegraphics[width=\linewidth]{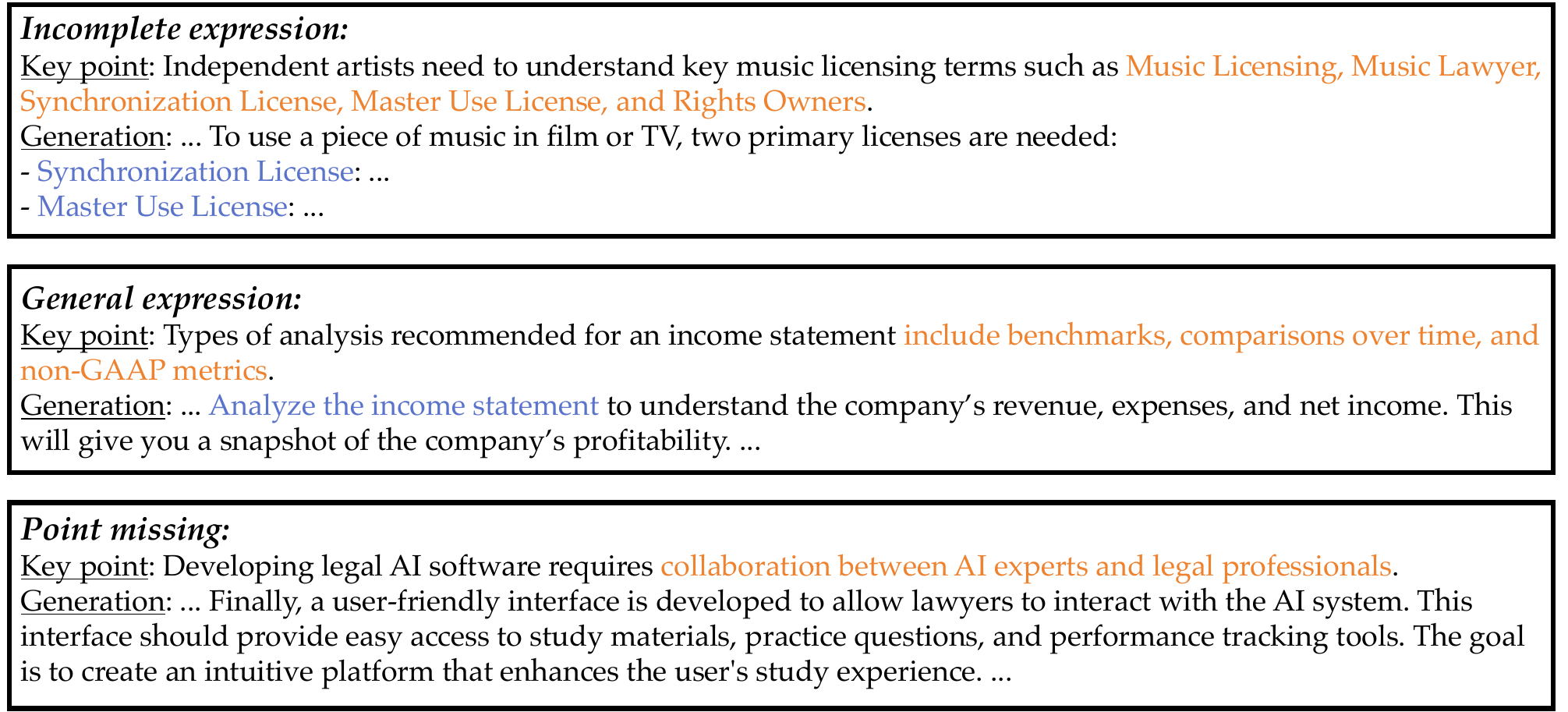}
    \caption{Cases where the model generation fails to include a key point. \textcolor[rgb]{0.93, 0.51, 0.18}{The key point} and \textcolor[rgb]{0.36, 0.45, 0.8}{the generation of LLMs} are highlighted.}
    \label{fig:error-analysis}
\end{figure*}

\noindent We provide insight on \textbf{RQ6} by plotting the relationship between LLMs' generation length and \metric{}, as in~\autoref{fig:gen-length-kpr}.
The figure reveals that for certain models, performance is nearly proportional to the length of the generated content. However, an exception to this pattern is observed with Qwen2-72B, suggesting that merely promoting lengthy generation is not the sole determinant of performance.

\subsection{Results of Different Evaluator}
\begin{table}[ht]
\setlength{\tabcolsep}{5pt} 
\fontsize{10}{10}\selectfont
\centering
\begin{threeparttable}
\begin{tabular}{lclc}
\toprule
\multicolumn{2}{c}{\textbf{GPT-4o}} & \multicolumn{2}{c}{\textbf{Llama3-70B}} \\ 
\cmidrule(lr){1-2} \cmidrule(lr){3-4} Model & KPR & Model & KPR\\
\midrule
GPT-4o & 0.579 & GPT-4o & 0.663 \\
Claude3-Sonnet & 0.477  & GPT-4-Turbo & 0.578  \\
GPT-4-Turbo & 0.469  & Claude-3-Sonnet & 0.568  \\
Qwen2-72B & 0.449  & Qwen2-72B & 0.562  \\
Phi-3-mini-128K  & 0.434  & Phi-3-mini-128K & 0.559  \\
Qwen2-7B & 0.416  & Qwen2-7B & 0.544  \\
Mixtral-8*22B & 0.383  & Mixtral-8*22B & 0.505  \\
Mistral-7B & 0.373  & Mistral-7B & 0.490  \\
Qwen2-1.5B & 0.262  & Qwen2-1.5B & 0.342 
\\ 
\bottomrule
\end{tabular}
\end{threeparttable}
\caption{\label{tab:evaluator}.The \metric for each model when using GPT-4o and Llama3-70B as evaluator}
\end{table}

To answer \textbf{RQ7}, We use GPT-4o and Llama3-70b as evaluators and evaluate the \metric{} of GPT-4o and Phi-3-mini-128K with both evaluators. The results shown in~\autoref{tab:evaluator} show that the model rankings remained largely consistent when evaluated by either GPT-4o or Llama3-70B, with only a minor position swap between Claude-3-Sonnet and GPT-4-Turbo. Llama3-70B generally assigned higher KPR scores due to its more lenient assessment of key points, while GPT-4o adopted a more conservative stance. These findings highlight the robustness of our comparative assessments.

\section{Analysis}
\label{sec:analysis}

In this section, we provide a qualitative analysis of the annotation and an analysis of the evaluator.

\subsection{Case Study on the Annotation}

We conduct a deeper analysis of why the LLMs fail to include key points in its generation. As shown in~\autoref{fig:error-analysis}, we identify \emph{three categories}: Incomplete expression, General expression, and Point missing. The incomplete expression category refers to instances where the generation partially includes the content of the key point but fails to cover it comprehensively. The general expression category typically occurs when a point is specific and detailed, yet the generation only vaguely or generally mentions it. The point missing category indicates cases where the generation completely omits a key point. In the first two categories, the generation still relatively incorporates information about the key point, whereas in the latter, the model fails to utilize key points to organize its generation. These three categories reflect the model's exploitation of key points during the generation process.

\subsection{Accuracy of the Evaluator}

To verify the reliability of GPT-4o as the \llmevaluator{}, 180 cases are sampled, comprising model generation, key point, and the assessment of GPT-4o (binary: entail or not). These cases are drawn from the models we evaluate. Two annotators are then required to validate these sampled cases. We find GPT-4o achieves a pleasant average accuracy rate of 87\%, demonstrating its reliability\footnote{This is also assured by setting the top-$p$ temperature$=0$ and fixed seed to ensure consistent evaluation results.} as the evaluator.
We report the IAA of the two annotators to be $\kappa = 0.61$ (Cohen's Kappa), indicating substantial agreement.

\subsection{Quality of the retrieved documents}

We conduct an additional experiment to assess document quality.
Specifically, we truncate the documents used for key point extraction into segments of 1024 tokens each. For each segment, we compute the embeddings of both the question and the segment using the BAAI/bge large-en-v1.5 model, and then calculate the cosine similarity between these embeddings.
The mean and standard deviation of the cosine similarity scores across all segments are 0.627 and 0.081, respectively. These results indicate that the documents utilized for extracting key points are of sufficiently high quality. The low standard deviation also suggests that the noises in those documents are low.

\subsection{Additional Evaluation Result of Key Point Precision}
\begin{table}[ht]
\setlength{\tabcolsep}{1pt} 
\fontsize{10}{10}\selectfont
\centering
\begin{threeparttable}
\begin{tabular}{>{\arraybackslash}p{1.8cm} 
                >{\centering\arraybackslash}p{0.8cm} 
                >{\centering\arraybackslash}p{0.8cm} 
                >{\centering\arraybackslash}p{0.8cm} 
                >{\centering\arraybackslash}p{2.7cm}}

\toprule
\textbf{Model} & \textbf{KPP} & \textbf{KPR} & \textbf{KPF} & \textbf{Response Length}  \\
\midrule
\textbf{GPT-4o} & 0.323 & 0.579 & 0.379 & 939.35 \\\textbf{Phi-3-mini} & 0.337 & 0.434 & 0.354 & 832.93 \\
\bottomrule
\end{tabular}
\end{threeparttable}
\caption{Evaluation results for key points in terms of precision and F1 score.}
\label{tab:kpp_kpf}
\end{table}

We incorporate two new metrics: Key Point Precision (\kpp{}) and Key Point F1-score (\kpf{}). \kpp measures the proportion of key points in the generated response that are actually present in the retrieved documents. \kpf combines \kpp and Key Point Recall (\metric{}) to provide a more comprehensive evaluation of key point accuracy.
We conduct an analysis of these metrics on 280 questions, comparing the performance of GPT-4o and Phi-3-mini-128K. The results are presented in~\autoref{tab:kpp_kpf}
From the result, it can observed that while GPT-4o has a longer average response length, it has a lower \kpp than Phi-3. Longer responses may not perform better across all metrics.
The lower \kpp for GPT-4o can be attributed to its tendency to incorporate parametric knowledge beyond the retrieved key points, demonstrating the depth of understanding but potentially reducing precision.
The \metric{} score provides a balanced view of performance, considering both recall and precision. This helps mitigate the bias towards longer responses that might be present in \metric{} alone.

\section{Conclusion}
\label{sec:conclusion}

This paper introduces a novel benchmark, \dataset, and a corresponding evaluation metric, \metric, to address the limitations of existing benchmarks for evaluating long-context and long-form RAG in LLMs. \dataset features intricate and practical questions with associated retrieved documents that faithfully replicate the characteristics of real-world RAG scenarios. \metric focuses on evaluating the LLM's ability to effectively exploit the retrieved documents by measuring the recall of key points extracted from these documents within the generated response. 
We conduct an evaluation on 9 LLMs using \dataset{} and \metric{}, presenting novel insights and analysis.

\section{Limitation}
\label{sec:limitation}

While the introduction of \dataset and \metric provides a significant step forward in evaluating long-context and long-form RAG, there are several limitations to this work. Firstly, the dataset, although diverse and carefully curated, is limited in size with only 280 questions. Expanding the dataset could provide more robust and generalizable insights. Secondly, the evaluation is primarily focused on English language content, which may not fully represent the capabilities of LLMs in handling other languages. Thirdly, the automated metric \metric, while innovative, may not perfectly capture the nuances of human evaluation, and its reliance on key point extraction could introduce additional biases or errors. Additionally, this study does not delve into the impact of different retrieval strategies on the performance of LLMs, which could be an important factor in RAG systems. Finally, the work primarily focuses on evaluating the effectiveness of LLMs in utilizing retrieved information, neglecting other aspects of long-form texts such as factual accuracy and logical coherence.

\section{Ethics Statement}
\label{sec:ethics}

This work focuses on ethical considerations in developing and evaluating LLMs for RAG. We curated the \dataset benchmark to minimize data contamination and reflect real-world challenges like low signal-to-noise ratio and dispersed information in retrieved documents. We ensured the dataset does not contain personally identifiable information or offensive content. The proposed evaluation metric, \metric, assesses the effectiveness of LLMs in leveraging retrieved information, providing insights into how models exploit external knowledge. By promoting transparency and responsible evaluation practices, this research aims to contribute to the development of accurate and ethically sound LLMs.

\section*{Acknowledgements}
The authors would like to thank the reviewers from the ACL Rolling Review for their thoughtful and constructive feedback. Their valuable insights have significantly enhanced the quality and clarity of our paper.
This work was supported by National Key Research and Development Program of China (2023YFC3304800).

\bibliography{custom}

\appendix
\section{Details about \dataset{}}

\subsection{Distribution of \dataset{}}
\label{sec: distribution}
\begin{figure*}[ht]
    \centering
    \includegraphics[width=\linewidth]{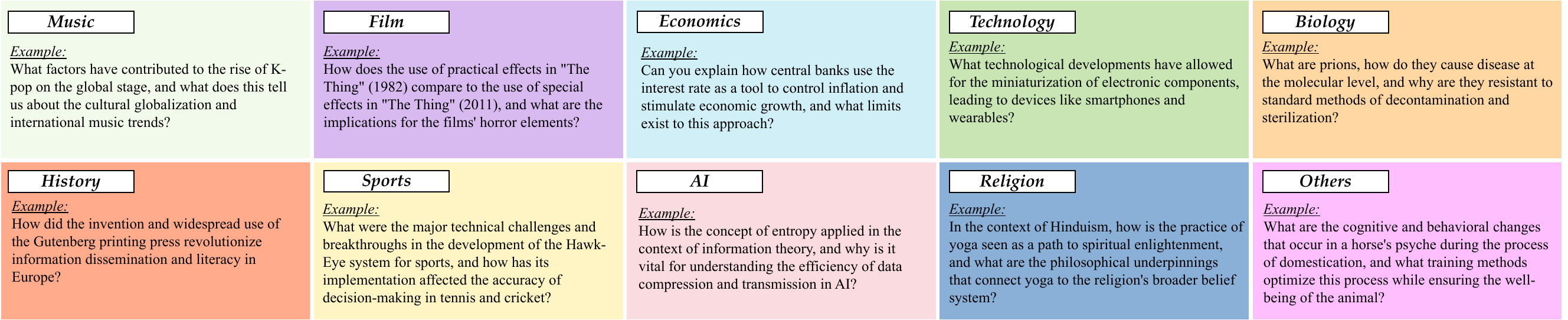}
    \caption{Examples of questions for each domain.}
    \label{fig:domain-cases}
\end{figure*}

\begin{figure}[ht]
    \centering
    \includegraphics[width=\linewidth]{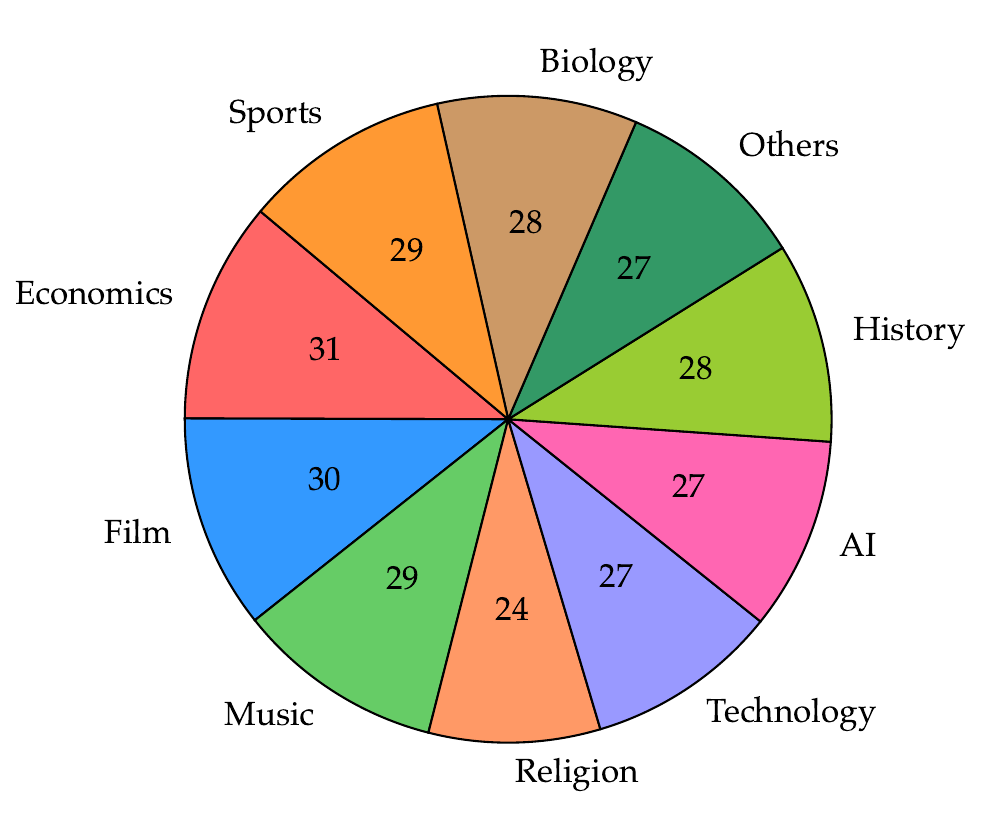}
    \caption{Domain distribution of questions in \dataset{}.}
    \label{fig:domain-distribution}
\end{figure}

\begin{figure}[ht]
    \centering
    \includegraphics[width=\linewidth]{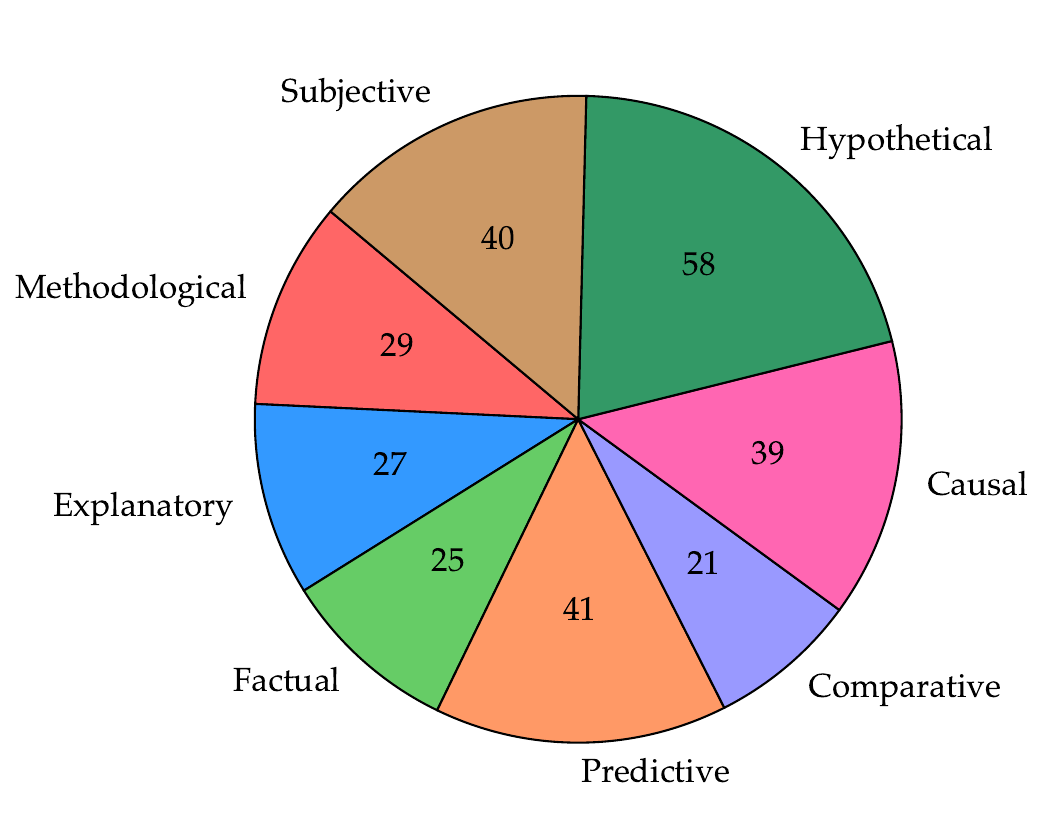}
    \caption{Question category distribution of questions in \dataset{}.}
    \label{fig:questiontype-distribution}
\end{figure}

\begin{figure}[ht]
    \centering
    \includegraphics[width=\linewidth]{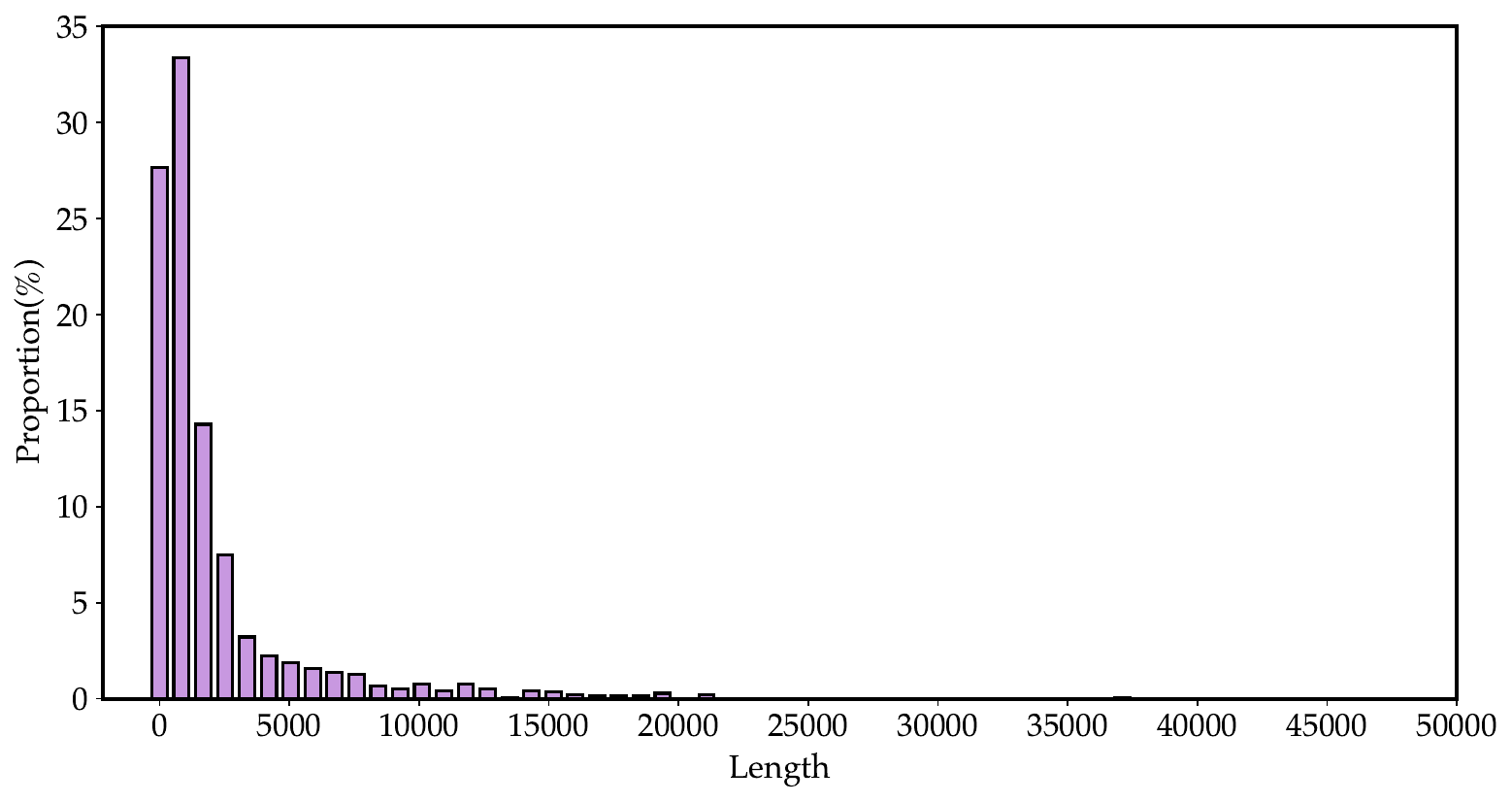}
    \caption{Length distribution of all retrieved documents. The horizontal coordinate represents the number of words in each document, while the vertical coordinate denotes the proportion.}
    \label{fig:length-distribution}
\end{figure}

We show the domain distribution and question category distribution of the questions in \dataset{} in~\autoref{fig:domain-distribution} and~\autoref{fig:questiontype-distribution}, respectively. It can be seen that both distributions are relatively uniform. We show one instance of each domain in~\autoref{fig:domain-cases}. Moreover, the length distribution of all retrieved documents is depicted in~\autoref{fig:length-distribution}. 

\subsection{The Score Function from AutoDS}
\label{sec: score_func}
In an effort to streamline the process of pre-filtering key points extracted by LLMs that do not adhere to our criteria, thereby reducing the workload of human annotators, we have implemented the scoring function available in AutoDS~\cite{zhang2024autonomous}. The rationale behind this approach is that LLMs often struggle with quantitative evaluation, particularly when it comes to assigning numerical scores that accurately reflect the importance of the extracted information~\cite{hopkins2023can, hu2023amortizing}. Nevertheless, DPO~\cite{rafailov2024direct} demonstrated that logits can be employed as a score function, and AutoDS also adopts this strategy.

\begin{figure}[ht]
    \centering
    \includegraphics[width=\linewidth]{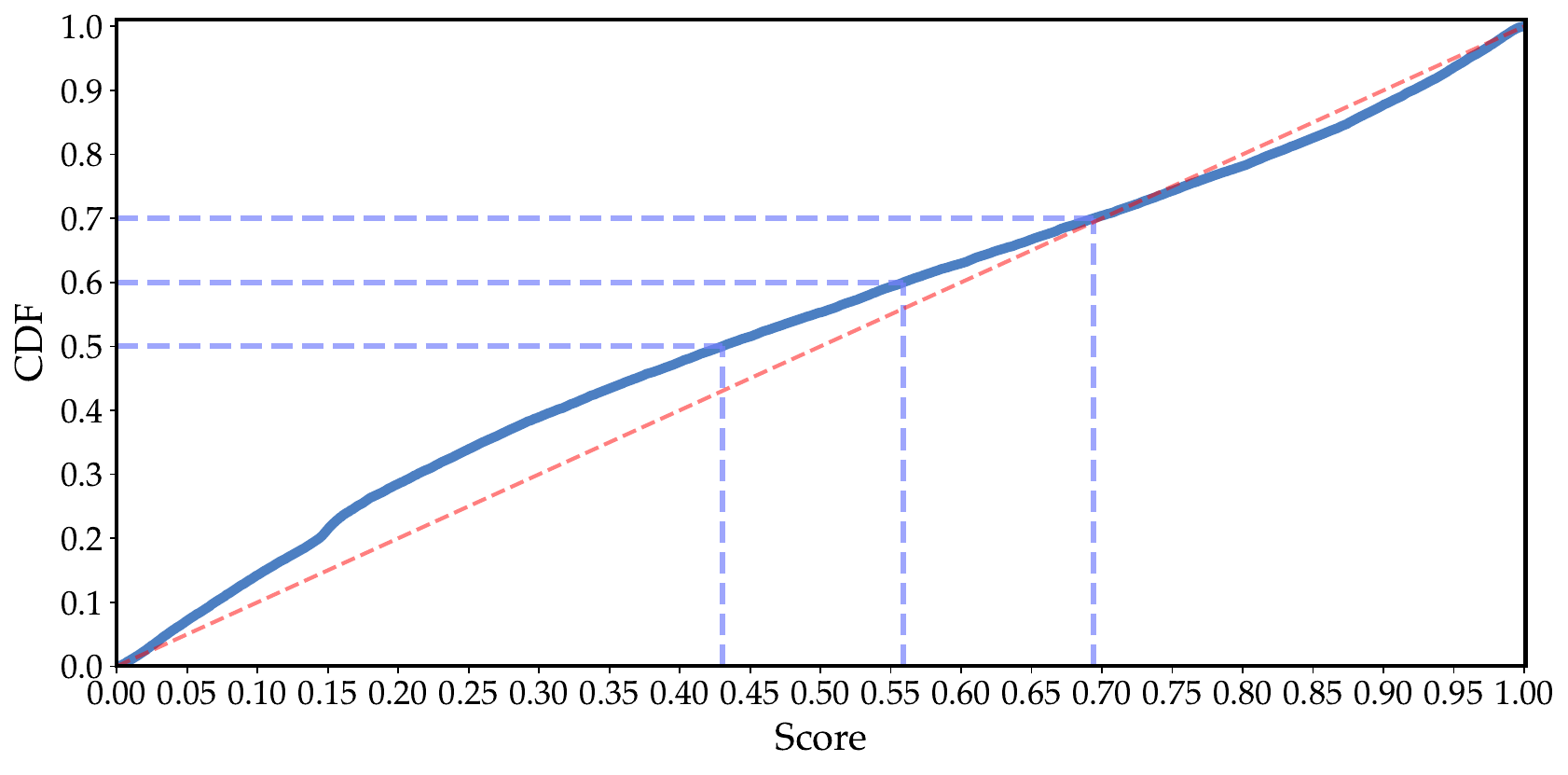}
    \caption{Score distribution of all points. The three dot lines in the plot indicate the scores when the CDF is 0.5, 0.6, and 0.7, respectively. The red dot line is a \(y=x\) diagonal. The closer to the red line,  the better the calibration.}
    \label{fig:score-distribution}
\end{figure}

This function evaluates the LLM’s propensity to agree or negate a claim in the content. With a carefully designed prompt, this function operates on the
logits corresponding to ``YES'' and ``NO'' responses to achieve content evaluation. The equation of the score function is:
\begin{equation}
\label{eq: score}
\fontsize{9}{9}\selectfont
    \text{Score}(\cdot)\!=\!\frac{\exp(\text{logit}(\text{``YES''}))}{\exp(\text{logit}(\text{``YES''}))\!+\!\exp(\text{logit}(\text{``NO''}))}
\end{equation}
In our work, we employ \href{https://llama.meta.com/llama3/}{LLama3-70b-Instruct} to compute the logits in~\autoref{eq: score}, utilizing the following prompt:
\begin{tcolorbox}[mybox, width=\linewidth,colframe=black!40, colback=white,title={\fontsize{10}{10}\selectfont Prompt for filtering the key points},coltitle=white,left=1pt,right=1pt,top=1pt,bottom=1pt] 
{
\texttt{%
\fontsize{9}{9}\selectfont
You are a helpful AI assistant. Your role is to evaluate whether a piece of information can serve as a key point in answering the question.\\
question: \{question\}\\
information: \{point\}\\
Can the above information directly help in addressing the question (thus being a key point)? You must respond with YES or NO.
}
}
\end{tcolorbox}
The score distribution of all the 28,611 points extracted is shown in~\autoref{fig:score-distribution}.
The fact that we can observe a near-diagonal distribution from the CDF (Cumulative Distribution Function) of~\autoref{fig:score-distribution} suggests~\autoref{eq: score} is well calibrated.
To determine an appropriate threshold, we randomly select 20 questions accompanied by 1,051 points and conduct annotation by two annotators. We compute the false positive rate for various threshold scores, which indicates the proportion of erroneously identified non-key points among the filtered points. The false positive rate of each individual's annotation results is averaged. We opt for a threshold of 0.7 due to its associated false positive rate being 8.7\%, thereby facilitating substantial data filtration while maintaining a tolerable error margin.

\subsection{Human Verification Details}

\begin{figure*}[htbp]
    \centering
    \includegraphics[width=\linewidth]{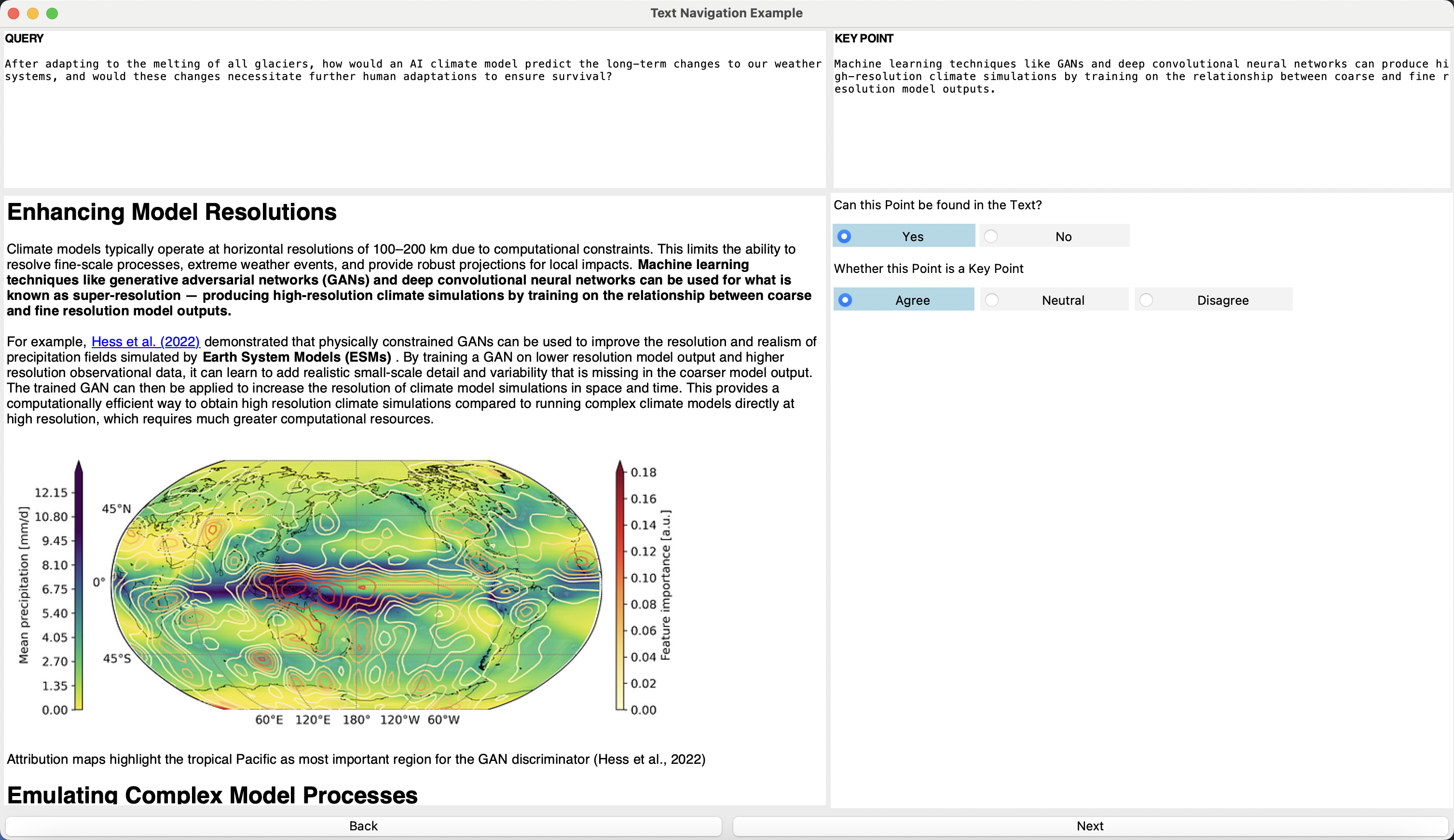}
    \caption{Our annotation interface. The interface comprises four utility areas. The upper left area displays the question, the lower left area presents the retrieved document, the upper right area represents the key point, and the lower right area is used for annotation.}
    \label{fig:annotation_ui}
\end{figure*}

\begin{figure*}[htbp]
    \centering
    \includegraphics[width=\linewidth]{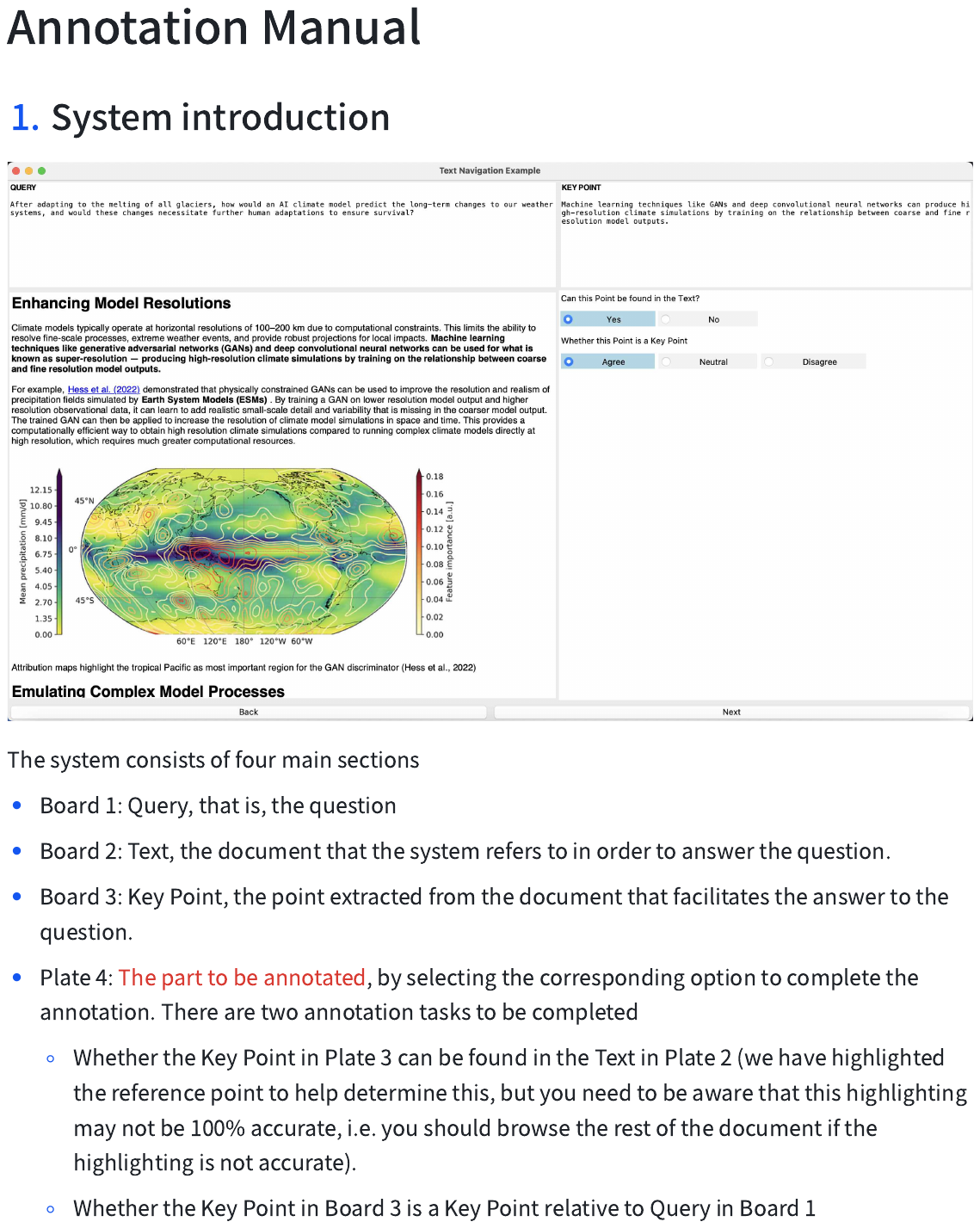}
    \caption{Annotation manual (Page 1 of 4).}
    \label{fig:annotation_handbook_1}
\end{figure*}

\begin{figure*}[htbp]
    \centering
    \includegraphics[width=\linewidth]{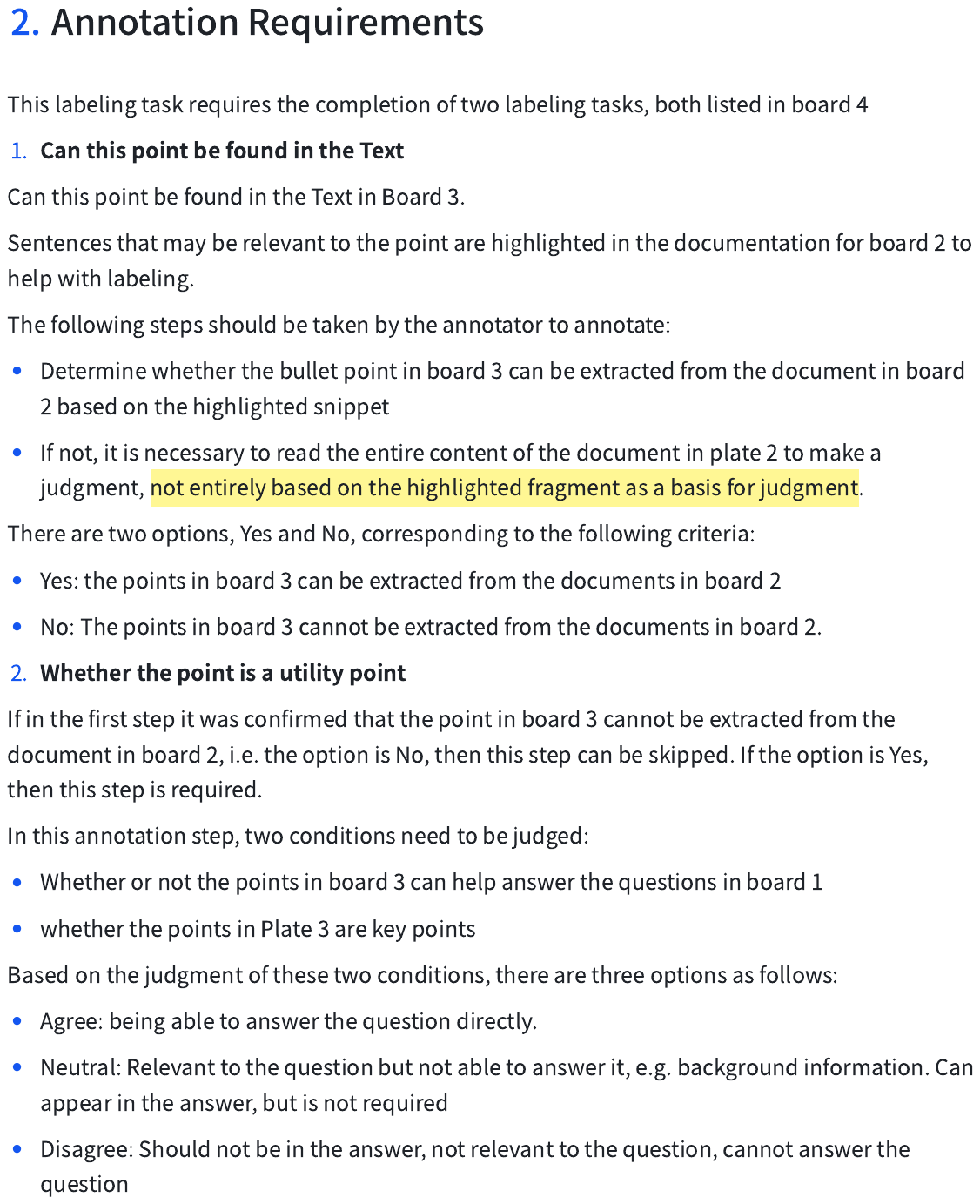}
    \caption{Annotation manual (Page 2 of 4).}
    \label{fig:annotation_handbook_2}
\end{figure*}

\begin{figure*}[htbp]
    \centering
    \includegraphics[width=\linewidth]{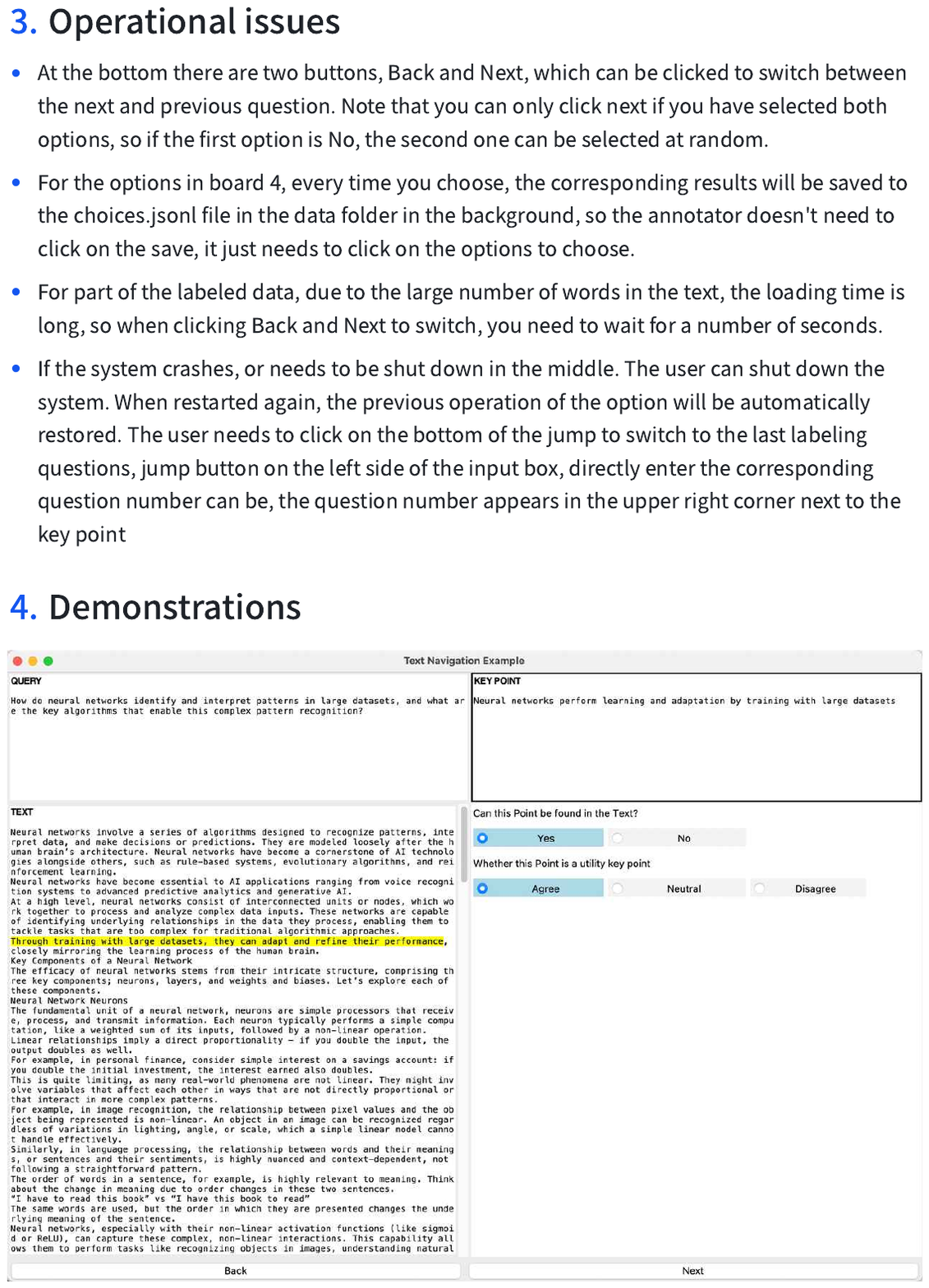}
    \caption{Annotation manual (Page 3 of 4).}
    \label{fig:annotation_handbook_3}
\end{figure*}

\begin{figure*}[htbp]
    \centering
    \includegraphics[width=\linewidth]{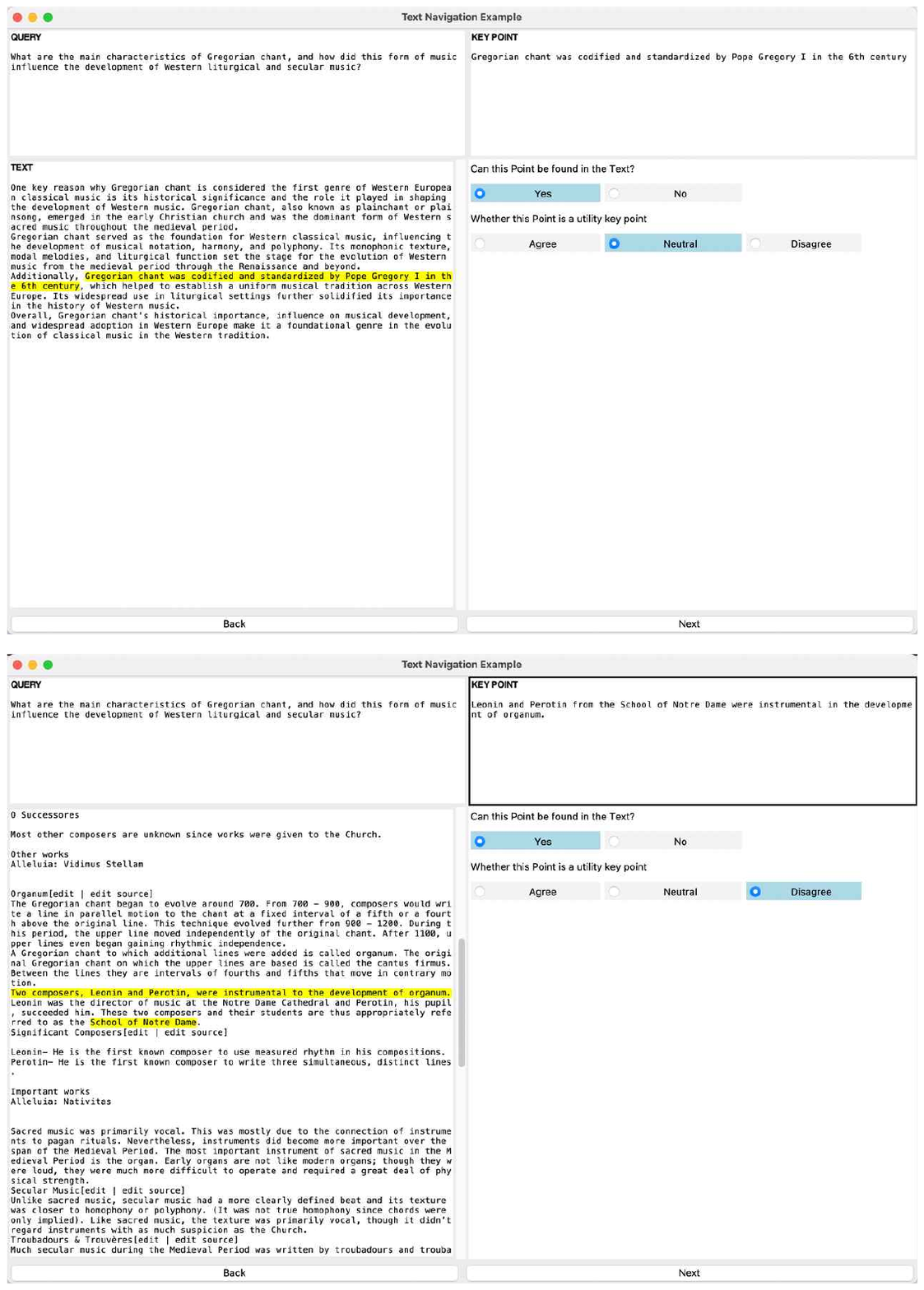}
    \caption{Annotation manual (Page 4 of 4).}
    \label{fig:annotation_handbook_4}
\end{figure*}

Four annotators are recruited to verify the key points extracted by the LLM. All the annotators are college students who major in English. The interface of the provided annotation tool is shown in~\autoref{fig:annotation_ui} alongside the annotation manual in~\autoref{fig:annotation_handbook_1}, \autoref{fig:annotation_handbook_2}, \autoref{fig:annotation_handbook_3} and \autoref{fig:annotation_handbook_4}.

This annotation interface contains the question, the retrieved document, and the extracted key point. The annotators are required to complete two tasks, including:
\begin{itemize}
    \item \textbf{Task 1 ``Can this Point be found in the Text?''}: Verifying whether this point is grounded by the document. 
    \item \textbf{Task 2 ``Whether this Point is a Key Point''}: Verifying whether this point is a key point. Three options are provided where ``Agree'' signifies the ability to directly address the question, ``Neutral'' indicates the provision of relevant background information related to the question without directly answering it, and ``Disagree'' denotes content that is not pertinent to the question and should be omitted.
\end{itemize}

We evenly distributed the key points derived from 560 questions among four annotators, with each annotator responsible for annotating key points associated with 140 questions. 

To ensure annotation reliability, we implemented a stringent quality control process.
For this process, 10\% of the key points were annotated by all four annotators. This overlapping subset served as a control group, allowing us to verify the consistency and quality of annotations across individuals.

Due to the quality constraints, we only selected the annotated data from two annotators to form our dataset, \dataset{}, which comprises 280 questions. From this dataset, we randomly sampled 200 key points that were jointly annotated by these two annotators. To evaluate the inter-annotator agreement, particularly in light of category imbalances, we utilized the free-marginal multi-rater kappa statistic~\cite{randolph2005free}. The kappa value of 0.56 indicates a moderate level of agreement among the annotators.

\section{Prompts Employed in Dataset Construction}

In the procedure of constructing \dataset{} with the automatic pipeline utilizing \llmdataset{}, we employ a series of prompts. 
Readers are expected to generate additional dataset samples in accordance with the guidelines outlined in~\autoref{sec:dataset} and by applying the following prompts for future research.

\label{sec: prompts-dataset}

\label{sec:prompts-dataset}

\noindent\textbf{Prompt for filtering questions in \eli{}:}
{
\texttt{%
\fontsize{9}{9}\selectfont
Please evaluate and score a given query based on the following criteria. Each satisfied criterion earns one point, with a maximum score of 5 points: \\
1. Exclude common sense questions: Filter out any queries that can be answered with basic common knowledge or a simple search engine query. Example: ``What is the boiling point of water?” or “What is the highest mountain in the world?'' \\
2. Ensure clarity: Filter out any queries that are unclear or have a vague scope. Example: ``Tell me something interesting.'' or ``Explain this.''\\
3. Ensure no ambiguity: Filter out any queries that could have multiple interpretations, ensuring the query has one clear answer or direction. Example: ``How do you do this?'' (needs to specify what ``this'' refers to) \\
4. Complexity requirement: Select queries that require deep thinking, detailed explanation, or multi-step reasoning. Example: ``How can graph neural networks be used to optimize recommendation algorithms in social networks?'' \\
5. Subjective opinion: Select queries that require the user to provide personal insights or subjective opinions. Example: ``What are your thoughts on the future of artificial intelligence in healthcare?'' \\
Please score the given query based on these criteria, with a range of 1 to 5 points. You should output the score wrapped with [], like [score]. \\
Here is the query: \\
{query} \\
Your score is: 
}

\noindent\textbf{Prompt for in-width evolving: }
{
\texttt{%
\fontsize{9}{9}\selectfont
You are a question generator. Your goal is to draw inspiration from the \#Given Question\# to create a brand new question. \\
This new question should belong to the same question category as the \#Given Question\#. But remember, the \#Created Question\# should be closely related to \{topic\} topic. \\
The LENGTH and difficulty level of the \#Created Question\# should be similar to that of the \#Given Question\#.\\
The \#Created Question\# must meet the following requirements:\\
1. The question should require a long-form response that includes several specific details.\\
2. Do not generate commonsense, vague, or ambiguous questions.\\
3. The question should be reasonable and can be responded to by humans.\\ 
4. You are encouraged to generate questions that require deep thinking, detailed explanations, or multi-step reasoning.\\
5. The \#Created Question\# should be very specific and niche within the topic of \{topic\}. The created question must be closely related to the topic of \{topic\}.\\
6. Follow the question styles in the \#Given Question\#. The \#Given Question\# belongs to type \{question\_type\}.\\
7. Wrap the \#Created Question\# in square brackets, like [created question]. You can generate multiple questions once a time and wrap each question in square brackets, like created question i: [question i].\\
\#Given Question\#:\\
\{seed\_question\}\\
\#Created Question\#:}

\noindent \textbf{Prompt for in-depth evolving:}
{
\texttt{%
\fontsize{9}{9}\selectfont
You are a question generator. Your goal is to draw inspiration from the \#Given Question\# to create a brand new question.
This new question should become the more complex version of the \#Given Question\# to make those famous AI systems (e.g., ChatGPT and GPT4) a bit harder to handle. But remember, the \#Created Question\# should be closely related to \{topic\} topic.\\
You can complicate the given prompt using the following methods: deepening, concretizing, increasing reasoning steps, and complicating input.\\
You should try your best not to make the \#Created Question\# become verbose, \#Created Question\# can only add 10 to 20 words into \#Given Question\#. \\
This new question should belong to the same question category as the \#Given Question\#.\\
The \#Created Question\# must meet the following requirements:\\
1. The question should require a long-form response that includes several specific details.\\
2. Do not generate commonsense, vague, or ambiguous questions.\\
3. The question should be reasonable and can be responded to by humans.\\ 
4. You are encouraged to generate questions that require deep thinking, detailed explanations, or multi-step reasoning.\\
5. The \#Created Question\# should be very specific and niche within the topic of \{topic\}. The created question must be closely related to the topic of \{topic\}.\\
6. Follow the question styles in the \#Given Question\#. The \#Given Question\# belongs to type 
\{question\_type\}.\\
7. Wrap the \#Created Question\# in square brackets, like [created question]. You can generate multiple questions once a time and wrap each question in square brackets, like created question i: [question i].\\
\#Given Question\#:\\
\{seed\_question\}\\
\#Created Question\#:
}

\noindent\textbf{Prompt for decomposing questions for retrieval:} 
\texttt{%
\fontsize{9}{9}\selectfont
You are a language analysis assistant capable of analyzing user questions to determine if and how to decompose the question.\\
**Problem Definition**\\
1. Simple question: Direct questions that only require simple information and do not include metaphors, multi-hop, multi-entity, or other complex logic questions;\\
2. Complex question: Includes multi-hop, metaphors, multi-entity, complex conditions, etc., requiring in-depth analysis and thought, as well as support from various types of information;\\
3. Vague question: Refers to questions with unclear intent, lacking a query subject, or ambiguously expressed content.\\
**Task Requirements**\\
1. For simple questions, there is no need to decompose the question;\\
2. For complex questions, decompose the complex question into multiple sub-questions based on the specific content of the question, ensuring the sub-question has a complete intent. Sub-questions need to be more concise and easier to search;\\
3. For vague questions, reasonably expand based on the information provided in the question, generating multiple related questions, each covering different aspects of the query subject.\\
Analyze user questions according to the given requirements and provide results. The results should be output in the format of an inline JSON.
Output format: \\
```json\\
{{``Question Type'': ``Simple/Complex/Vague'', ``Sub-questions'': [``...'']}}\\
```\\
Question: \{query\}
}

\noindent\textbf{Prompt for extracting key points:}
{
\texttt{%
\fontsize{9}{9}\selectfont
Based on the text provided, identify key points in the text that directly help in responding to the query. \\
Format your response as follows: each point should start with ``Point [number]:'', followed by its content and spans in the text that entails the key point. \\
IMPORTANT: The output must be of the format `` Point [number]: <point\_start>[content of point] \\<point\_end><span\_start>[span1]<span\_end><span\_s\\ tart>[span2]<span\_end>'' \\
IMPORTANT: Ensure each point is helpful in responding to the query. Keep the point using the original language and do not add explanations.\\
IMPORTANT: Each span must be a single consecutive verbatim span from the corresponding passages. Copy verbatim the spans, don't modify any word! \\
Here is an example: \\
<One-shot Demonstration> \\
Remember: \\
- key points can be abstracted or summarized, but the span must be a copy of the original text. The content of the key point does NOT need to be the same as that of the span.\\
- These key points must be helpful in responding to the query.\\
- Copy verbatim the spans, don't modify any word! If there are multiple spans for a point, separate them with <span\_start> and <span\_end> tokens. \\
{[Query]}: \{query\} \\
{[Text]}: \{text\} \\
{[Key Point]}
}

\noindent\textbf{Prompt for entailment identification: }
{
\texttt{%
\fontsize{9}{9}\selectfont
Your task is to determine whether a claim is entailed with a document.\\
You should evaluate whether the information in the document supports or describes the claim provided.\\
You must provide an answer based on whether the document does entail the claim, does not entail the claim, or is neutral.\\
If the claim is entailed, you should provide snippets from the document that support this claim.\\
Document: \{document\}\\
Claim: \{claim\}\\
Provide your answer as [yes], [no], or [neutral]. State your reason behind the answer and provide snippets from the document that support this claim if your answer is [yes].
}

\noindent\textbf{Prompt for key point deduplication and aggregation:}
{
\texttt{%
\fontsize{9}{9}\selectfont
Based on the points extracted from a piece of text, identify and remove any duplicate points to streamline the list.\\
REMEMBER:\\
- The de-duplicated points need to contain all the original points.\\
- An original point cannot exist in two different de-duplicated points at the same time.\\
- Format your response as follows: each unique point should start with ``Point [number]:'', followed by its content and corresponding original point numbers. The corresponding original point numbers should be wrapped with [].\\
Here is an example:\\
<One-shot Demonstration>\\
Remember to not add any new points, only de-duplicate the existing ones. And do not add any explanations.\\
{[Original Points]}\\
\{points\}\\
{[De-duplicate Points]}
}

\end{document}